%% file: main.tex
\documentclass[10pt,twocolumn,letterpaper]{article}

\usepackage[margin=0.75in]{geometry}
\usepackage[T1]{fontenc}
\usepackage[utf8]{inputenc}
\usepackage{times}
\usepackage{microtype}
\usepackage{xspace}
\usepackage{graphicx}
\usepackage{booktabs}
\usepackage{tabularx}
\usepackage{multirow}
\usepackage{amsmath}
\usepackage{amssymb}
\usepackage{bbm}
\usepackage{xcolor}
\usepackage{pifont}
\usepackage{algorithm}
\usepackage{algpseudocode}
\usepackage{listings}

\usepackage[numbers,sort&compress]{natbib}
\usepackage{dblfloatfix}
\usepackage[colorlinks=true,allcolors=blue,breaklinks=true]{hyperref}

\makeatletter
\g@addto@macro\UrlBreaks{\do\/\do-\do\_\do\.\do\?\do\&\do\=\do\#}
\makeatother
\newcommand{\ind}{\mathbbm{1}}

\title{MultiBind: A Benchmark for Attribute Misbinding in \\Multi-Subject Generation}

\author{
Wenqing Tian$^{1,2,4,*}$, Hanyi Mao$^{3,4,*}$, Zhaocheng Liu$^{4,\dagger}$, Lihua Zhang$^{4}$,\\
Qiang Liu$^{1,2}$, Jian Wu$^{4}$, Liang Wang$^{1,2,\dagger}$\\[0.5em]
\normalsize $^{1}$New Laboratory of Pattern Recognition, Institute of Automation, Chinese Academy of Sciences\\
\normalsize $^{2}$School of Artificial Intelligence, University of Chinese Academy of Sciences\\
\normalsize $^{3}$The University of Chicago\\
\normalsize $^{4}$ByteDance\\[0.25em]
\normalsize $^{*}$Equal contribution\qquad
\normalsize $^{\dagger}$Corresponding authors: \texttt{lio.h.zen@gmail.com}, \texttt{wangliang@nlpr.ia.ac.cn}
}
\date{}

\begin{document}
\maketitle

\input{sections/abstract}
\input{sections/intro}
\input{sections/related}
\input{sections/dataset}
\input{sections/metrics}
\input{sections/results}
\input{sections/conclusion}

\begingroup
\sloppy
\setlength{\emergencystretch}{2em}
\bibliographystyle{unsrtnat}
\bibliography{main}
\endgroup

\appendix
\input{sections/supplementary}

\end{document}

%% file: sections/abstract.tex
\begin{abstract}
Subject-driven image generation is increasingly expected to support fine-grained control over multiple entities within a single image. In multi-reference workflows, users may provide several subject images, a background reference, and long, entity-indexed prompts to control multiple people within one scene. In this setting, a key failure mode is \emph{cross-subject attribute misbinding}: attributes are preserved, edited, or transferred to the wrong subject. Existing benchmarks and metrics largely emphasize holistic fidelity or per-subject self-similarity, making such failures hard to diagnose. We introduce \textbf{\textsc{MultiBind}}, a benchmark built from real multi-person photographs. Each instance provides slot-ordered subject crops with masks and bounding boxes, canonicalized subject references, an inpainted background reference, and a dense entity-indexed prompt derived from structured annotations. We also propose a dimension-wise confusion evaluation protocol that matches generated subjects to ground-truth slots and measures slot-to-slot similarity using specialists for face identity, appearance, pose, and expression. By subtracting the corresponding ground-truth similarity matrices, our method separates self-degradation from true cross-subject interference and exposes interpretable failure patterns such as drift, swap, dominance, and blending. Experiments on modern multi-reference generators show that \textsc{MultiBind} reveals binding failures that conventional reconstruction metrics miss.

\end{abstract}

%% file: sections/intro.tex
\section{Introduction}
Multi-reference image generation has rapidly evolved into a practical workflow where users rely on fine-grained, entity-indexed prompts to independently control multiple subjects in editing, design, and content creation
\cite{radford2021clip,li2023blip2,liu2023llava,ramesh2022dalle2,rombach2022ldm,brooks2023instructpix2pix,ruiz2023dreambooth,ye2023ipadapter,liu2023cones2,wang2025msdiffusion,wu2025uno,chen2025multiref,oshima2025multibanana}.
In many real-world use cases, users provide several subject reference images and write explicit ``Subject~A/B/C'' blocks to specify distinct attributes, actions, and relations for each entity within a single scene
\cite{openai2023dalle3systemcard,midjourney2026characterref,adobe2026referenceimage,qian2025layercomposer}.
As these systems become more capable, a central question emerges for the community:
how do we reliably measure fine-grained controllability under long, structured instructions in multi-subject settings?

In this paper, we focus on \emph{multi-reference, multi-subject} image generation. Users provide several subject reference images, often alongside a background reference, and a long, entity-indexed prompt that assigns different attributes, actions, and relations to specific subjects in a shared scene.
The goal is not merely to generate a globally plausible image, but to compose all subjects into one scene while preserving the identity and unspecified attributes of each reference, simultaneously binding the requested edits to the correct target subjects.

\begin{figure*}[t]
  \centering
  \includegraphics[width=0.98\textwidth]{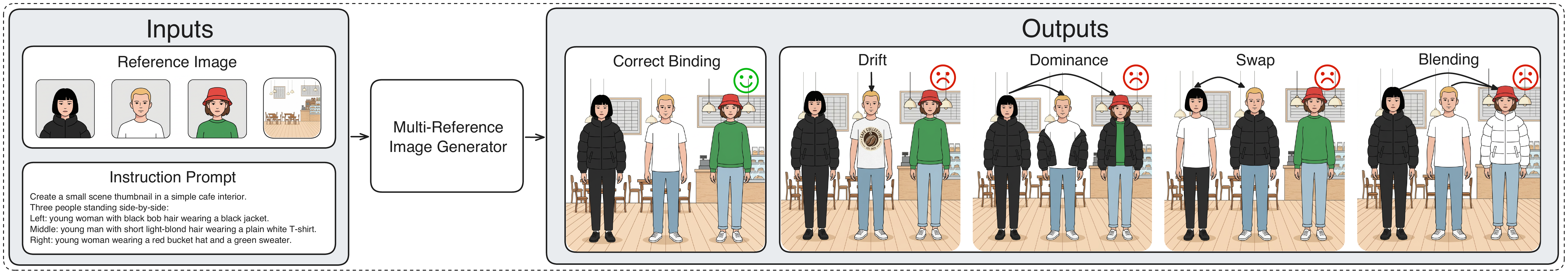}
    \caption{Multi-reference, multi-subject generation takes multiple subject references, a background reference, and an entity-indexed prompt as input. Correct binding respects slot-specific instructions, while failures include but are not limited to \textbf{drift} (a subject deviates from its own target without being confused with other subjects), \textbf{dominance} (one reference dominates multiple subjects), \textbf{swap} between subjects, or \textbf{blending} where attributes are mixed across subjects. This diagram illustrates the patterns through variations in clothing across different subjects.}
  \label{fig:teaser}
\end{figure*}

This binding requirement is exactly where current systems often fail.
When per-subject controls become detailed and intertwined, visual cues and textual directives can leak across subjects: a jacket intended for Subject~A appears on Subject~B, a smile lands on the wrong face, or apparel cues are averaged across references.
We refer to this failure mode as \emph{cross-subject attribute misbinding}. Closely related to the ``binding'' and ``leakage'' errors studied in compositional generation
\cite{dahary2024boundedattention,trusca2024objectattribute,jang2024mudi}, this failure mode yields outputs that may look globally coherent at a glance while still violating specific user intent.
In other words, each subject must satisfy two requirements at once: follow the requested edits, and retain unspecified attributes from its \emph{own} reference without absorbing cues from others.
Evaluation in this setting therefore has to be both subject-specific and attribute-specific.
Fig.~\ref{fig:teaser} illustrates the setting and several representative failure modes, including drift, dominance, swap, and blending across subjects.

However, existing evaluation protocols lag behind these practical capabilities.
Many works and benchmarks emphasize global similarity signals, such as CLIP-based alignment and distributional fidelity
\cite{radford2021clip,hessel2021clipscore,heusel2018ganstrainedtimescaleupdate}.
Some personalization methods explicitly study subject confusion, but quantitative evaluation is typically limited to face-identity self-similarity or pairwise matching in an identity embedding space
\cite{ruiz2023dreambooth,ye2023ipadapter,li2023blipdiffusion,li2023photomaker,zong2024easyref,wang2025msdiffusion,wu2025uno,xiao2023fastcomposer,jang2024mudi,chen2025xverse,she2025mosaic,xu2025contextgen}.
While such scalars may correlate with overall fidelity, they provide weak diagnostics for complex controllability:
they cannot answer \emph{who confuses with whom}, nor can they provide quantitative indicators to distinguish generic self-degradation (drift) from cross-subject interference
\cite{xiao2023fastcomposer,jang2024mudi,chen2025xverse,she2025mosaic,xu2025contextgen}.
Recent VLM-based protocols can improve overall prompt adherence assessment
\cite{kamath2025geneval2},
but they are largely reference-free and do not expose per-subject correspondence errors.


A second limitation concerns benchmark construction, particularly the choice of target images. Some benchmarks synthesize targets by prompting a generator to compose multiple subjects into a single image \cite{saha2025sigmagen}. While scalable, generating targets from synthetic prompts creates an inherent dilemma. Because these scenes are unanchored from real images, complex prompts risks generating internally inconsistent images, whereas overly simple ones fail to rigorously test multi-subject control. Furthermore, other benchmarks do not provide a paired ground-truth target image at all, relying instead on reference-free (often VLM-based) judging for scoring \cite{oshima2025multibanana,kamath2025geneval2}. These issues motivate our shift toward benchmarks grounded in real targets, which naturally guarantee both rich, consistent details and explicit correspondence supervision.

We introduce \textbf{\textsc{MultiBind}}, a benchmark designed to stress-test long-prompt, multi-subject controllability with explicit subject correspondence supervision.
Each instance is anchored to a unique real target image, and provides:
(i) per-subject ground-truth crops with instance masks and bounding boxes,
(ii) canonicalized subject reference images and an inpainted background reference, and
(iii) structured attribute descriptions compiled into a long, entity-indexed prompt.
Grounding each condition in a real target enables attribute-rich prompts that remain realistic, diverse, and internally consistent, while also supporting reproducible similarity-based scoring.

Beyond the benchmark, we propose a confusion-aware evaluation protocol that makes subject-attribute misbinding directly measurable.
For each attribute dimension, we first compute subject-to-subject similarity matrices between generated subjects and ground-truth subjects using dimension-specific specialists.
To isolate the changes introduced by generation, we subtract the inherent similarities between ground-truth subjects to compute baseline-corrected delta matrices.
This step effectively disentangles generic quality degradation (a subject losing its own features, reflected on the matrix diagonal) from cross-subject interference (a subject absorbing features from others, reflected off-diagonal).
The resulting diagnostics expose whether failures manifest as drift, specific swaps, dominant confusers, or blending.
Fig.~\ref{fig:main} provides an overview of the full \textsc{MultiBind} pipeline.

\paragraph{Contributions.}
Our main contributions are summarized as follows:

\textbf{Benchmark.} We establish \textsc{MultiBind}, a robust benchmark for multi-subject and multi-reference generation. Unlike previous datasets, it is grounded in real target images and provides exhaustive annotations, including per-subject masks, bounding boxes, and background references, alongside structured captions rewritten into entity-indexed prompts.

\textbf{Evaluation.} We introduce a dimension-wise evaluation protocol that leverages specialist representations to produce confusion-aware similarity and delta matrices. This framework enables precise diagnostics of common multi-subject failure modes, such as identity drift, subject swapping, and attribute blending.

\textbf{Analysis.} Through a systematic evaluation under the \textsc{MultiBind} regime, we benchmark state-of-the-art multi-reference generators and report fine-grained binding trends, offering new insights into how models represent and reason with multiple logical entities.

%% file: sections/related.tex
\section{Related Work}
\label{sec:related}

\paragraph{Subject-driven image generation.}
Subject-driven and reference-conditioned generation aims to preserve subject identity and appearance from one or more reference images while following new textual instructions.
Early personalization approaches adapt diffusion models per subject via fine-tuning or token learning. This enables identity preservation but requires per-subject optimization
\cite{ruiz2023dreambooth}.
To bypass test-time tuning, recent works inject image conditions through lightweight adapters or specialized modules, improving usability for image-guided prompting and editing
\cite{ye2023ipadapter,li2023photomaker}.
Extending these capabilities to multi-subject settings exposes a critical failure mode: multiple high-fidelity references often interfere, causing swaps and attribute bleeding.
A range of methods attempt to mitigate this via localized attention or layout guidance \cite{xiao2023fastcomposer,liu2023cones2}, alongside recent multi-subject personalization pipelines
\cite{wang2025msdiffusion,jang2024mudi,wu2025uno,she2025mosaic,chen2025xverse,xu2025contextgen}.
While these methods advance generation capabilities, their evaluation commonly remains coarse---such as measuring only diagonal similarity to each subject's own reference. This limitation motivates a benchmark capable of explicitly diagnosing cross-subject interference.

\paragraph{Multi-subject benchmarking.}
Several benchmarks have begun to stress-test multi-reference composition at scale.
For instance, MRBench evaluates group image references \cite{zong2024easyref},
MultiRef-bench targets controllable generation with multiple visual anchors \cite{chen2025multiref},
and MultiBanana systematically varies reference-set conditions to probe robustness \cite{oshima2025multibanana}.
Other works release paired datasets alongside generation methods (e.g., XVerseBench, MS-Bench, LAMICBench++, and IMIG-100K)
\cite{chen2025xverse,wang2025msdiffusion,xu2025contextgen}.
Additionally, specialized evaluations focus on multi-human identity preservation \cite{borse2025multihuman} or multi-image context generation \cite{wu2026miconbench}.
These efforts provide valuable coverage of prompt and reference-set diversity. However, many settings still rely on LLM- or VLM-as-a-judge scoring or weak supervision. This makes the results sensitive to the choice of evaluator and susceptible to benchmark drift as these models evolve \cite{kamath2025geneval2}.
More importantly, diagnosing \emph{who} interferes with \emph{whom} requires explicit correspondence supervision. This demands paired targets with deterministic, slot-indexed entity correspondences (such as instance masks or bounding boxes) and specific per-entity attributes.
Without such grounding, evaluation frequently reduces to judge-based scoring or simple diagonal preservation, failing to quantify off-diagonal confusion like swaps and attribute leakage across subjects.
\textsc{MultiBind} is designed for this setting by pairing each multi-reference condition with a unique \emph{real} ground-truth target and explicit slot-level supervision, enabling reproducible, confusion-aware diagnostics.

\begin{table*}[t]
  \centering
  \scriptsize
  \setlength{\tabcolsep}{2.5pt}
  \renewcommand{\arraystretch}{1.06}
  \caption{Comparison with related benchmarks and data resources.
  $\checkmark$ indicates explicit support; $\triangle$ indicates partial support via structured task specifications (e.g., prompt templates, image placeholders, preset bounding boxes, layout images, or task-level constraints) without full entity-level grounding; \textcolor{red}{\ding{55}} indicates not provided or not the focus; ``--'' indicates not applicable. For ``Paired target'', we report the source of the target image when available.}
  \label{tab:benchmark_comparison}
  \begin{tabular}{lccccc}
    \toprule
    \shortstack{Benchmark/Resource} &
    \shortstack{Multi\\ref.} &
    \shortstack{Multi\\subj.} &
    \shortstack{Paired\\target} &
    \shortstack{Fine-grained \\ Entity-level prompt} &
    \shortstack{Misbinding\\diagnosis} \\
    \midrule
    \multicolumn{6}{l}{\textbf{Benchmarks}} \\
    MRBench~\cite{zong2024easyref} & $\checkmark$ & \textcolor{red}{\ding{55}} & Real & \textcolor{red}{\ding{55}} & \textcolor{red}{\ding{55}} \\
    MultiRef-bench~\cite{chen2025multiref} & $\checkmark$ & \textcolor{red}{\ding{55}} & \shortstack{Real+Synth} & $\triangle$ & \textcolor{red}{\ding{55}} \\
    MultiBanana~\cite{oshima2025multibanana} & $\checkmark$ & $\checkmark$ & \textcolor{red}{\ding{55}} & \textcolor{red}{\ding{55}} & \textcolor{red}{\ding{55}} \\
    XVerseBench~\cite{chen2025xverse} & $\checkmark$ & $\checkmark$ & \textcolor{red}{\ding{55}} & \textcolor{red}{\ding{55}} & \textcolor{red}{\ding{55}} \\
    MS-Bench~\cite{wang2025msdiffusion} & $\checkmark$ & $\checkmark$ & \textcolor{red}{\ding{55}} & $\triangle$ & \textcolor{red}{\ding{55}} \\
    LAMICBench++~\cite{xu2025contextgen} & $\checkmark$ & $\checkmark$ & \textcolor{red}{\ding{55}} & $\triangle$ & \textcolor{red}{\ding{55}} \\
    MultiHuman-Testbench~\cite{borse2025multihuman} & $\checkmark$ & $\checkmark$ & \textcolor{red}{\ding{55}} & \textcolor{red}{\ding{55}} & \textcolor{red}{\ding{55}} \\
    MICON-Bench~\cite{wu2026miconbench} & $\checkmark$ & $\checkmark$ & \textcolor{red}{\ding{55}} & $\triangle$ & \textcolor{red}{\ding{55}} \\
    \midrule
    \multicolumn{6}{l}{\textbf{Data resources}} \\
    IMIG-100K~\cite{xu2025contextgen} & $\checkmark$ & $\checkmark$ & Synth & $\checkmark$ & -- \\
    SIGMA-SET27K~\cite{saha2025sigmagen} & $\checkmark$ & $\checkmark$ & Synth & $\checkmark$ & -- \\
    \midrule
    \textbf{MultiBind (ours)} & $\checkmark$ & $\checkmark$ & Real & $\checkmark$ & $\checkmark$ \\
    \bottomrule
  \end{tabular}
\end{table*}

\paragraph{Attribute binding and diagnostic evaluation.}
Binding failures are widely studied in text-only compositional generation, where models misassociate entities and modifiers (e.g., ``a pink sunflower and a yellow flamingo'').
For example, SynGen improves attribute correspondence by aligning cross-attention maps according to syntactic structure \cite{rassin2023linguisticbinding}.
In parallel, fine-grained text-to-image evaluation has progressed beyond global alignment using object- or question-based checks \cite{ghosh2023geneval,hu2023tifa}.
However, these works do not address multi-reference interference, where the dominant failure mode is not merely incorrect text grounding, but cross-subject confusion among multiple visual anchors.
In multi-subject personalization, evaluation frequently reports diagonal identity preservation (often face-focused) or holistic image similarity. As discussed, these scalars cannot reveal whether a failure is caused by generic self-degradation (drift) or by cross-subject interference (confusion).
While methods like MuDI target identity decoupling and report multi-subject diagnostics \cite{jang2024mudi}, existing protocols remain limited in attributing interference across \emph{multiple} attribute dimensions (such as clothing, pose, and expression) under a unified framework.
Our protocol addresses this limitation by employing dimension-wise specialists and converting continuous similarities into calibrated binary indicators.
This yields interpretable confusion matrices and baseline-corrected metrics for specific failure patterns---including drift, dominance, swaps, and blending---under strict ground-truth supervision.

%% file: sections/dataset.tex
\section{The \textsc{MultiBind} Dataset}
\label{sec:dataset}

\subsection{Task Definition}
\textsc{MultiBind} instantiates multi-reference generation as a \emph{real-image reconstruction} task: given per-subject reference images, a background reference, and an entity-indexed prompt, the model must reconstruct a \emph{real} ground-truth target image $I_{\mathrm{gt}}$.
We use real images as targets because they exhibit diverse, fine-grained controllable factors while remaining globally coherent.

We focus exclusively on \emph{human} subjects. Multi-person generation is a common and particularly challenging use case for subject misbinding. It also offers relatively well-defined semantic dimensions, making failures more directly measurable and comparable across models.

Assuming $I_{\mathrm{gt}}$ contains $N$ subject slots, we formalize the visual factors as $\mathcal{Z}(I_{\mathrm{gt}}) = (\{s_i\}_{i=1}^{N}, b, R, E)$, where $b$, $R$, and $E$ denote background, relations, and environment factors respectively. Each subject $s_i = (s_i^{\text{edit}}, s_i^{\text{preserve}})$ is partitioned into two sets. The \emph{edit} set $s_i^{\text{edit}} = \{\text{pose}, \text{expression}\}$ contains attributes altered in the canonicalized references that must be recovered via prompt guidance. The \emph{preserve} set $s_i^{\text{preserve}} = \{\text{identity}, \text{appearance}\}$ contains dimensions that must carry over strictly from the reference image without leaking across slots.

\begin{figure*}[!htbp]
\centering
\includegraphics[width=1.0\textwidth]{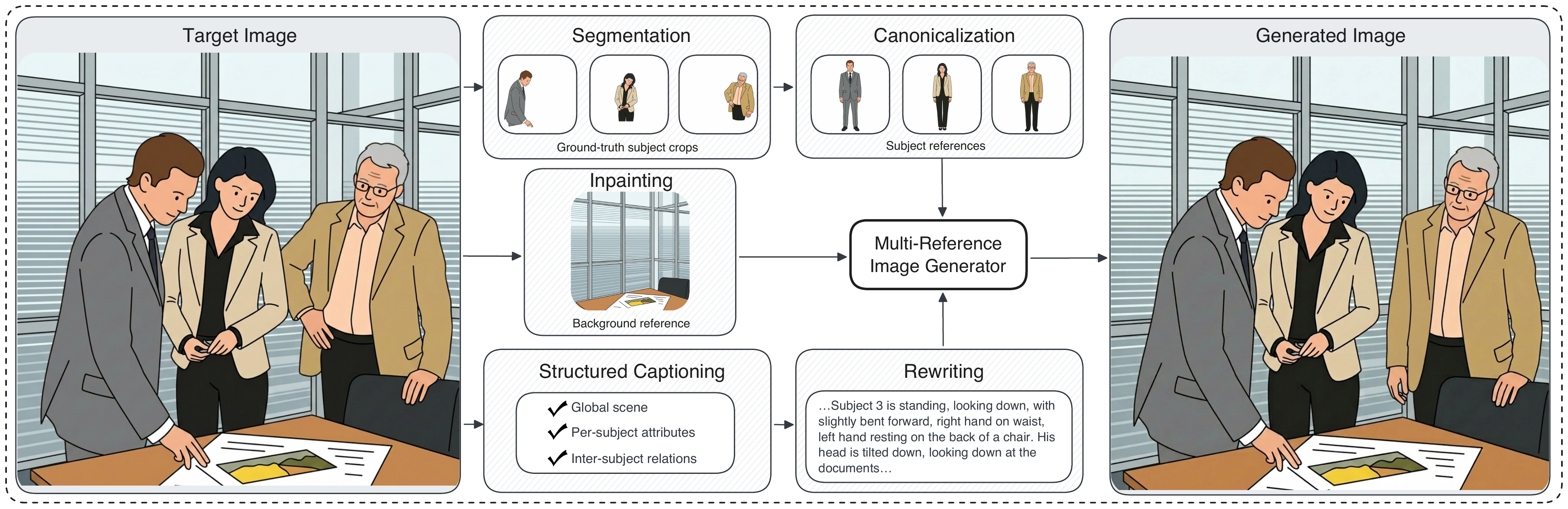}
\caption{Starting from a real target image $I_{\mathrm{gt}}$, we segment subjects to obtain ground-truth crops, canonicalize each subject to build per-subject reference images, and inpaint the removed regions to produce a background reference.
We then perform structured captioning and rewrite the resulting fields into a long, entity-indexed prompt.
A multi-reference generator conditions on the subject references, background reference, and prompt to produce a synthesized image $I_{\mathrm{gen}}$ intended to reconstruct $I_{\mathrm{gt}}$.}
\label{fig:main}
\end{figure*}

Given the condition $C = (\{r_i^{\mathrm{subject}}\}_{i=1}^{N}, r^{\mathrm{background}}, p)$, where $r_i^{\mathrm{subject}}$ are standardized subject references, $r^{\mathrm{background}}$ is the background reference, and $p$ is the entity-indexed prompt, a generator $G$ produces $I_{\mathrm{gen}} = G(C)$ to reconstruct $I_{\mathrm{gt}}$ with correct subject-attribute binding (Fig.~\ref{fig:main}).

\subsection{MultiBind Instance Construction and Statistics}
\label{sec:dataset_images}

Starting from a real target image $I_{\mathrm{gt}}$, we construct one canonicalized subject reference image $r_i^{\mathrm{subject}}$ per slot, an inpainted background reference $r^{\mathrm{background}}$, and compile structured annotations into the entity-indexed prompt $p$ (Fig.~\ref{fig:main}).
The full automated and manual pipeline, including instance segmentation, generative canonicalization and inpainting, strict multi-stage quality control, and rule-based prompt rewriting, is detailed in the supplementary material.

\paragraph{Dataset sources and statistics.}
We curate \textsc{MultiBind} from four public datasets: CIHP~\cite{gong2018pgn}, LV-MHP-v2~\cite{li2017mhp}, Objects365~\cite{shao2019objects365}, and COCO~\cite{lin2014coco}. \textsc{MultiBind} contains 508 instances and 1,527 human subjects. The dataset features 118, 269, and 121 instances with two, three, and four subjects, respectively. Every instance utilizes an entity-indexed prompt (referencing fixed slots like ``Subject~A'') with an average length of 474 words. Detailed dataset distributions are provided in the supplementary material.

%% file: sections/metrics.tex
\providecommand{\suppContinuousMetricsRef}{the supplementary material}

\section{\textsc{MultiBind} Evaluation}
\label{sec:evaluation}

\textsc{MultiBind} evaluates cross-subject binding in three steps.
For each ground-truth target image $I_{\mathrm{gt}}$ and a generated reconstruction $I_{\mathrm{gen}}$,
we (1) extract person instances from $I_{\mathrm{gen}}$ and match them to the $N$ ground-truth subject slots, obtaining the set of successfully matched slots $\mathcal{M}\subseteq\{1,\ldots,N\}$;
(2) compute dimension-wise subject-to-subject similarity matrices using dimension-specific specialists; and
(3) derive confusion-oriented diagnostics from baseline-corrected similarity deltas.
We discuss the details of the matching algorithm in the supplementary material, and report the successful match count and mean IoU in Sec.~\ref{sec:results_main}.
Note that different models may match different subsets of subjects for the same target instance.
To ensure a fair comparison, every model is evaluated on the same subject subset for a given instance.

\subsection{Dimension-wise similarity matrices}
\label{sec:eval_similarity}

This section defines the per-dimension similarity matrices that serve as the common input to all subsequent confusion analyses.

\paragraph{Per-slot crops and specialists.}
Consider one instance.
For each matched slot $i\in\mathcal{M}$, the dataset provides the ground-truth subject crop $o_i^{\mathrm{gt}}$ (Sec.~\ref{sec:dataset_images}).
We also extract the corresponding generated crop $o_i^{\mathrm{gen}}$ from $I_{\mathrm{gen}}$ using the matched mask.
We evaluate four attribute dimensions
\(\mathcal{D}=\{\text{face identity},\\  \text{appearance},\ \text{pose},\ \text{expression}\}\)
(Table~\ref{tab:eval_specialists}).
For each dimension $d\in\mathcal{D}$, we compute specialist features
\(f_i^{x,(d)} \,=\, g_d(o_i^{x})\) for $x\in\{\mathrm{gt},\mathrm{gen}\}$,
and compare slots with a dimension-appropriate similarity $\mathrm{sim}_d(\cdot,\cdot)$.

\begin{table}[t]
\centering
\footnotesize
\caption{Dimension-specific specialists used in \textsc{MultiBind} evaluation.}
\label{tab:eval_specialists}
\begin{tabular}{@{}llc@{}}
\toprule
Dimension $d$ & Specialist $g_d$ & Similarity $\mathrm{sim}_d$ \\
\midrule
Face identity & InsightFace (Buffalo\_L) \cite{deng2019arcface,insightface} & cosine \\
Pose & ViTPose \cite{xu2022vitpose} & OKS \\
Appearance & Qwen3-VL-Emb.-8B \cite{li2026qwen3vlembedding} & cosine \\
Expression & Qwen3-VL-Emb.-8B & cosine \\
\bottomrule
\end{tabular}
\end{table}

\paragraph{Valid slots.}
Some specialists are only defined when the required visual evidence is present
(e.g., the face specialist requires a detected face).
For each dimension $d$, let $\mathcal{V}^{(d)}\subseteq\{1,\ldots,N\}$ denote the set of ground-truth slots where the specialist output is valid.
All per-subject evaluations are performed on the row index set
\(\mathcal{I}^{(d)} \,=\, \mathcal{M}\cap \mathcal{V}^{(d)}\),
so rows correspond to \emph{matched generated subjects} with valid specialist outputs, while columns always range over \emph{valid ground-truth subjects} $j\in\mathcal{V}^{(d)}$.

\paragraph{Similarity matrices.}
For each dimension $d$, we build two similarity matrices of shape $|\mathcal{I}^{(d)}|\times |\mathcal{V}^{(d)}|$:
\begin{equation}
\begin{aligned}
S_{\mathrm{gt}}^{(d)}[i,j]
&= \mathrm{sim}_d\!\big(f_{i}^{\mathrm{gt},(d)},\ f_{j}^{\mathrm{gt},(d)}\big),\\
S_{\mathrm{gen}}^{(d)}[i,j]
&= \mathrm{sim}_d\!\big(f_{i}^{\mathrm{gen},(d)},\ f_{j}^{\mathrm{gt},(d)}\big),
\qquad i\in\mathcal{I}^{(d)},\ j\in\mathcal{V}^{(d)}.
\end{aligned}
\label{eq:sim_matrices_joint}
\end{equation}

All subsequent confusion analyses operate on the baseline-corrected \emph{delta matrix}
\begin{equation}
\Delta^{(d)} \,=\, S_{\mathrm{gen}}^{(d)} - S_{\mathrm{gt}}^{(d)}.
\label{eq:delta_matrix}
\end{equation}

The key role of $S_{\mathrm{gt}}^{(d)}$ is to provide an \emph{instance-specific baseline}:
its off-diagonal entries quantify how similar the \emph{ground-truth} subjects already are to each other in dimension $d$.
Subtracting this baseline isolates the change introduced by generation.
Concretely:
(i) the diagonal $\Delta^{(d)}[i,i]$ measures \emph{self-retention} (how close the generated subject in slot $i$ stays to its own ground-truth subject);
and (ii) an off-diagonal entry $\Delta^{(d)}[i,j]\ (j\neq i)$ becomes positive when the generated subject in slot $i$ moves \emph{toward} ground-truth subject $j$ \emph{beyond} what is already implied by the ground-truth similarity between subjects $i$ and $j$.
We report aggregated diagonal and off-diagonal values in \suppContinuousMetricsRef{}.

\subsection{Binary indicators and failure patterns}
\label{sec:eval_patterns}

To provide interpretable diagnostics for subject- and image-level failure modes, we binarize $\Delta^{(d)}$ into
(a) a diagonal \emph{self-consistency} signal and (b) an off-diagonal \emph{cross-subject confusion} signal,
using thresholds calibrated to human annotations.
Specifically, for each matched generated subject crop $o_{i}^{\mathrm{gen}}$ and each dimension $d$,
human labelers annotate whether it is (1) consistent with $o_{i}^{\mathrm{gt}}$ and (2) confused with $o_{j}^{\mathrm{gt}},\ j\neq i$ in dimension $d$.
The thresholds for consistency $\mathrm{thresh}_{\mathrm{cons}}^{(d)}$ and confusion $\mathrm{thresh}_{\mathrm{conf}}^{(d)}$
are derived by maximizing the F1 score between $\Delta^{(d)}[i,i]$ and consistency labels,
and between $\Delta^{(d)}[i,j]$ and confusion labels, respectively.
Annotation details and threshold values are reported in the supplementary material.

\paragraph{Binary matrices.}
Using the calibrated thresholds, we define two binary matrices for each dimension $d$:
\begin{align}
\mathrm{Cons}^{(d)}[i,j]
&= \ind\Big[(j=i)\ \wedge\ \big(\Delta^{(d)}[i,i]\ge \mathrm{thresh}_{\mathrm{cons}}^{(d)}\big)\Big],
\label{eq:bin_cons}\\
\mathrm{Conf}^{(d)}[i,j]
&= \ind\Big[(j\neq i)\ \wedge\ \big(\Delta^{(d)}[i,j]\ge \mathrm{thresh}_{\mathrm{conf}}^{(d)}\big)\Big],
\label{eq:bin_conf}
\end{align}
where $i\in\mathcal{I}^{(d)}$ and $j\in\mathcal{V}^{(d)}$.
Here $\mathrm{Cons}^{(d)}$ marks self-consistent \emph{diagonal} matches.
We call any off-diagonal pair $(i,j)$ with $\mathrm{Conf}^{(d)}[i,j]=1$ a \emph{confusion link} (also called a ``confusion edge'' in a graph view):
it indicates that, in dimension $d$, the generated subject assigned to slot $i$ is anomalously close to the \emph{wrong} ground-truth subject $j$.

\paragraph{Subject-level outcomes.}
For each generated subject (row) $i\in\mathcal{I}^{(d)}$, define
\begin{align}
\mathrm{Confused}_i^{(d)}
&= \ind\Big[\bigvee_{\substack{j\in\mathcal{V}^{(d)}, j\neq i}} \mathrm{Conf}^{(d)}[i,j]\Big],
\label{eq:confused_row}\\
\mathrm{Inconsistent}_i^{(d)}
&= \neg\mathrm{Cons}^{(d)}[i,i],
\label{eq:inconsistent_row}\\
\mathrm{Success}_i^{(d)}
&= \mathrm{Cons}^{(d)}[i,i]\wedge\neg\mathrm{Confused}_i^{(d)},
\label{eq:success_row}\\
\mathrm{Drift}_i^{(d)}
&= \mathrm{Inconsistent}_i^{(d)}\wedge\neg\mathrm{Confused}_i^{(d)}.
\label{eq:drift_row}
\end{align}
Dataset-level subject rates are computed by averaging over the subjects of all instances.

\paragraph{Image-level failure patterns.}
Define the combined match indicator matrix
\begin{equation}
M^{(d)}[i,j] \,=\, \mathrm{Cons}^{(d)}[i,j]\ \lor\ \mathrm{Conf}^{(d)}[i,j],
\label{eq:M_or}
\end{equation}
so that $M^{(d)} \in \{0,1\}^{|\mathcal{I}^{(d)}|\times |\mathcal{V}^{(d)}|}$.
Let
\begin{align}
r_i^{(d)} &= \|M^{(d)}[i,:]\|_1,\
c_j^{(d)} = \|M^{(d)}[:,j]\|_1,\\
n_{\mathrm{conf}}^{(d)} &= \sum_{i\in\mathcal{I}^{(d)}}\sum_{\substack{j\in\mathcal{V}^{(d)}\\ j\neq i}}\mathrm{Conf}^{(d)}[i,j],
\label{eq:conf_edge_count}
\end{align}
which are the row and column degrees of $M^{(d)}$ and the total number of off-diagonal confusion links.
From them we detect three structured patterns, reported as image-level rates:
\begin{align}
\ind_{\mathrm{swap}}^{(d)}
\!&= \ind\left[
\begin{aligned}
n_{\mathrm{conf}}^{(d)} &> 0 \\
\wedge\ \max_{i\in\mathcal{I}^{(d)}} r_i^{(d)} &\le 1 \\
\wedge\ \max_{j\in\mathcal{V}^{(d)}} c_j^{(d)} &\le 1
\end{aligned}
\right],
\label{eq:swap_pat}\\
\ind_{\mathrm{dom}}^{(d)}
&= \ind\Big[\exists!\ j\in\mathcal{V}^{(d)}\ \text{s.t.}\ c_j^{(d)} = |\mathcal{I}^{(d)}|\Big],
\label{eq:dom_pat}\\
\ind_{\mathrm{blend}}^{(d)}
&= \ind\Big[\max_{i\in\mathcal{I}^{(d)}} r_i^{(d)} \ge 2\Big].
\label{eq:blend_pat}
\end{align}
Intuitively, \emph{swap} corresponds to a permutation-like assignment with at least one off-diagonal confusion link,
\emph{dominance} to a column-wise collapse onto a single ground-truth subject,
and \emph{blending} to a row-wise match to multiple ground-truth subjects.
Note that these patterns are defined heuristically, and one could define other indicators as needed based on the same binary matrices.

\subsection{Global pattern shift: row-wise JS}
\label{sec:eval_js}

To summarize how each row distribution changes (including probability mass moving off the diagonal), we compute a row-wise Jensen--Shannon (JS) shift.
Define the row distribution induced by similarities
\begin{equation}
\begin{aligned}
q^{(d)}_{x}(i,j)
&= \frac{\exp\big(S^{(d)}_{x}[i,j]\big)}{\sum_{k\in\mathcal{V}^{(d)}}\exp\big(S^{(d)}_{x}[i,k]\big)},
\\ & j\in\mathcal{V}^{(d)},\ x\in\{\mathrm{gt},\mathrm{gen}\}.
\end{aligned}
\end{equation}
We then report
\begin{equation}
\mathrm{JS}^{(d)}
\,=\,
\frac{1}{|\mathcal{I}^{(d)}|}\sum_{i\in\mathcal{I}^{(d)}}
\mathrm{JS}\!\Big(q_{\mathrm{gt}}^{(d)}(i,\cdot)\ \big\|\ q_{\mathrm{gen}}^{(d)}(i,\cdot)\Big).
\label{eq:js_row}
\end{equation}

\newcommand{\MultiBindContinuousMetricsAppendix}{
\subsection{Continuous confusion metrics}
\label{app:continuous_metrics}
\label{sec:eval_metrics}

For completeness, we also derive threshold-free scalar summaries from $\Delta^{(d)}$ that separate diagonal self degradation from off-diagonal cross-subject mixing.
These metrics are useful for measuring the magnitude of the change before thresholding and for checking that the main conclusions are not artifacts of the calibrated binarization.

\paragraph{Diagonal degradation.}
\begin{equation}
D_{\mathrm{self}}^{(d)}
\,=\,
-\frac{1}{|\mathcal{I}^{(d)}|}
\sum_{i\in\mathcal{I}^{(d)}} \Delta^{(d)}[i,i].
\label{eq:self_degradation}
\end{equation}

\paragraph{Off-diagonal mixing.}
Let $\delta_{i,j}^{(d)} = [\Delta^{(d)}[i,j]]_+$ for $j\neq i$.
We summarize diffuse and concentrated mixing via
\begin{align}
C_{\mathrm{mean}}^{(d)}
&= \frac{1}{|\mathcal{I}^{(d)}|}
\sum_{i\in\mathcal{I}^{(d)}}
\frac{1}{|\mathcal{V}^{(d)}|-1}
\sum_{\substack{j\in\mathcal{V}^{(d)}\\ j\neq i}}
\delta_{i,j}^{(d)},
\label{eq:mean_confusion}\\
C_{\mathrm{worst}}^{(d)}
&= \frac{1}{|\mathcal{I}^{(d)}|}
\sum_{i\in\mathcal{I}^{(d)}}
\max_{\substack{j\in\mathcal{V}^{(d)}\\ j\neq i}}
\delta_{i,j}^{(d)}.
\label{eq:worst_confusion}
\end{align}
}

%% file: sections/results.tex
\providecommand{\suppGenerationProtocolRef}{the supplementary material}
\providecommand{\suppContinuousResultsRef}{the supplementary material}
\providecommand{\suppCaseRef}{the supplementary material}

\section{Experiments}
\label{sec:results}

\subsection{Experimental Setup}
\label{sec:results_setup}

\subsubsection{Models}
We evaluate six image generation systems: three closed-source models, Gemini 3 Pro Image (Nano Banana Pro)~\cite{google2025gemini3proimage_modelcard}, GPT-Image-1.5~\cite{gpt}, and Seedream 4.5~\cite{seedream}; and three open-source models, HunyuanImage-3.0-Instruct~\cite{hunyuan}, Qwen-Image-Edit-2511~\cite{qwen}, and OmniGen2~\cite{omnigen2}.

We do not include several recent open-source multi-subject reference methods
(e.g., \cite{chen2025xverse,she2025mosaic,wu2025uno,wang2025msdiffusion,xu2025contextgen})
because most rely on CLIP-style text encoders with short context windows (commonly 77 tokens)
or limited-context T5-style encoders (e.g., 512 tokens), which are insufficient for our long, entity-indexed prompts.

\subsubsection{Multi-reference image generation}
For each \textsc{MultiBind} instance, models are conditioned on subject references $\{r_i^{\mathrm{subject}}\}_{i=1}^{N}$,
a background reference $r^{\mathrm{background}}$, and the fine-grained, entity-indexed prompt $p$ (Fig.~\ref{fig:teaser}).
We standardize output resolution across models and fix inference settings whenever the interface allows it.
The shared reconstruction setup, together with the model-specific settings explicitly reported there, is given in \suppGenerationProtocolRef.

\subsubsection{Metrics}
We report two complementary sets of metrics.

\textbf{Holistic reconstruction metrics} compare each generated image with the real target:
FID$\downarrow$ \cite{heusel2018ganstrainedtimescaleupdate} (distribution-level fidelity),
CLIP-I$\uparrow$ \cite{radford2021clip} and DINO$\uparrow$ \cite{caron2021dino} (image-level similarity),
and AES$\uparrow$, a pretrained aesthetic predictor score that summarizes overall visual appeal\cite{discus0434_aesthetic_predictor_v2_5}.
These metrics capture overall reconstruction quality, but they are not designed to isolate binding failures.

\textbf{Binding diagnostics} are computed from the subject-level similarity matrices after slot matching.
First, we use the row-wise Jensen--Shannon shift $\mathrm{JS}^{(d)}\downarrow$ (Sec.~\ref{sec:eval_js}), which measures how each subject's distribution over candidate ground-truth slots changes under generation.
Second, we report the rates of subject- and image-level diagnostics derived from binarized self-consistency and confusion indicators: \textbf{Success}, \textbf{Confused}, \textbf{Inconsistent}, \textbf{Drift}, and the structured patterns \textbf{Swap}, \textbf{Dominance}, and \textbf{Blending}, as discussed in Sec.~\ref{sec:eval_patterns}. We also report the aggregated raw values of diagonal degradation and off-diagonal mixing as continuous counterparts in \suppContinuousResultsRef.

\subsection{Holistic reconstruction and overall binding shift}
\label{sec:results_main}

Table~\ref{tab:main_results} summarizes the holistic metrics.
We additionally report the number of matched subject slots and the mean IoU between ground-truth and generated subjects to reflect each model's ability to produce the correct number of subjects in approximately correct locations.

\begin{table*}[t]
\centering
\small
\caption{Holistic metrics on \textsc{MultiBind}. The "Matched" metric reports the number of aligned subject slots (out of $1,305$), where the total reflects the subject count of the intersection of images successfully generated by all evaluated models.}
\label{tab:main_results}
\begin{tabular}{lccccccc}
\toprule
Model & FID$\downarrow$ & CLIP-I$\uparrow$ & DINO$\uparrow$ & AES$\uparrow$ & JS$\downarrow$ & Matched$\uparrow$ & Mean IoU$\uparrow$\\
\midrule
Nano Banana Pro & \textbf{80.51} & \textbf{0.87} & \textbf{0.81} & 5.06 & \textbf{0.0075} & \textbf{1294} & 0.40 \\
GPT-Image-1.5 & 82.95 & 0.84 & 0.77 & \textbf{5.44} & \underline{0.0086} & \textbf{1294} & \underline{0.40} \\
Seedream 4.5 & 81.52 & \underline{0.85} & \underline{0.77} & 4.92 & 0.0103 & 1282 & 0.34 \\
Hunyuan-Image-3.0-Instruct & \underline{81.23} & 0.80 & 0.76 & \underline{5.06} & 0.0129 & \underline{1287} & \textbf{0.42} \\
Qwen-Image-Edit-2511 & 94.26 & 0.76 & 0.56 & 4.59 & 0.0160 & 1080 & 0.31 \\
OmniGen2 & 94.65 & 0.77 & 0.66 & 4.65 & 0.0166 & 1038 & 0.32 \\
\bottomrule
\end{tabular}
\end{table*}

The closed-source models outperform the three open-source baselines on the holistic metrics.
Among all models, Nano Banana Pro gives the strongest overall reconstruction, achieving the best FID, CLIP-I, DINO, and global JS.
GPT-Image-1.5 attains the highest AES and ties for the largest number of matched subject slots, indicating strong visual quality together with reliable subject count and placement.
Hunyuan-Image-3.0-Instruct achieves the best Mean IoU, but its higher JS shows that good localization does not necessarily translate into stable subject binding.
Qwen-Image-Edit-2511 and OmniGen2 trail on both holistic similarity and slot matching.

\subsection{Failure-pattern diagnosis}
\label{sec:results_patterns}

Table~\ref{tab:pattern_rates} reports thresholded subject-level and image-level rates derived from the binary self-consistency and confusion indicators of Sec.~\ref{sec:eval_patterns}.
These metrics directly indicate whether a subject stays aligned with its own slot, drifts away from it, or becomes confused with another subject, making them our main diagnostic view.

\begin{table*}[!t]
\centering
\small
\caption{Subject- and image-level rates (\%). Success, Confused, Inconsistent, and Drift are subject-level rates; Swap, Dominance, and Blending are image-level pattern rates. Abbrev.: Suc.=Success, Conf.=Confused, Inc.=Inconsistent, Dom.=Dominance, Bld.=Blending. HunyuanImage-3 denotes HunyuanImage-3-Instruct; Qwen-Image-Edit denotes Qwen-Image-Edit-2511.}
\label{tab:pattern_rates}
\begin{tabular}{@{}llcccc@{}ccc@{}}
\toprule
Dim & Model &
\multicolumn{4}{c}{Subject-level (\%)} &
\multicolumn{3}{c}{Image-level (\%)} \\
\cmidrule(lr){3-6}\cmidrule(lr){7-9}
& &
\multicolumn{1}{c}{Suc.$\uparrow$} &
\multicolumn{1}{c}{Conf.$\downarrow$} &
\multicolumn{1}{c}{Inc.$\downarrow$} &
\multicolumn{1}{c}{Drift$\downarrow$} &
\multicolumn{1}{c}{Swap$\downarrow$} &
\multicolumn{1}{c}{Dom.$\downarrow$} &
\multicolumn{1}{c}{Bld.$\downarrow$} \\
\midrule

\multirow{6}{*}{Face}
& \mbox{Nano Banana Pro} & \textbf{84.0} & \underline{13.1} & \textbf{5.8} & \textbf{2.8} & \underline{2.0} & \textbf{3.7} & 25.3 \\
& \mbox{GPT-Image-1.5}   & \underline{82.3} & \textbf{12.8} & \underline{7.8} & \underline{4.9} & \textbf{1.5} & \underline{3.9} & \underline{24.4} \\
& \mbox{Seedream 4.5}    & 58.6 & 36.1 & 12.1 & 5.3 & 3.2 & 14.5 & 53.7 \\
& \mbox{HunyuanImage-3}  & 38.7 & 15.6 & 56.3 & 45.6 & 11.6 & \textbf{3.7} & \textbf{15.8} \\
& \mbox{Qwen-Image-Edit} & 43.7 & 36.2 & 43.8 & 20.1 & 17.1 & 11.7 & 31.5 \\
& \mbox{OmniGen2}        & 41.9 & 30.7 & 47.6 & 27.5 & 18.2 & 16.0 & 26.5 \\
\midrule

\multirow{6}{*}{Appearance}
& \mbox{Nano Banana Pro}& \textbf{95.4} & \textbf{3.4} & \textbf{3.7} & \underline{1.2} & \textbf{2.0} & \textbf{0.5} & \textbf{3.4} \\
& \mbox{GPT-Image-1.5}   & \underline{94.5} & \underline{4.7} & \underline{4.0} & \textbf{0.8} & \underline{2.5} & \underline{0.7} & \underline{5.4} \\
& \mbox{Seedream 4.5}    & 91.8 & 5.7 & 6.7 & 2.5 & \underline{2.5} & 0.9 & 6.6 \\
& \mbox{HunyuanImage-3}  & 78.0 & 16.2 & 15.0 & 5.8 & 2.9 & 1.6 & 21.2 \\
& \mbox{Qwen-Image-Edit} & 48.4 & 37.5 & 46.5 & 14.1 & 22.9 & 8.4 & 24.9 \\
& \mbox{OmniGen2}        & 57.5 & 34.5 & 36.5 & 8.0 & 18.9 & 12.8 & 23.7 \\
\midrule

\multirow{6}{*}{Pose}
& \mbox{Nano Banana Pro}& \textbf{68.0} & \textbf{1.2} & \textbf{31.8} & \textbf{30.8} & \textbf{0.4} & \textbf{2.2} & \textbf{0.9} \\
& \mbox{GPT-Image-1.5}   & \underline{60.7} & \underline{1.9} & 38.6 & 37.5 & \underline{0.5} & 7.1 & \underline{1.9} \\
& \mbox{Seedream 4.5}    & \underline{60.7} & 2.5 & \underline{38.0} & \underline{36.8} & 1.8 & 7.2 & 3.2 \\
& \mbox{HunyuanImage-3}  & 59.1 & 2.5 & 40.0 & 38.4 & \underline{0.5} & \underline{4.5} & 2.7 \\
& \mbox{Qwen-Image-Edit} & 40.4 & 5.5 & 57.7 & 54.1 & 3.6 & 9.3 & 4.1 \\
& \mbox{OmniGen2}        & 52.7 & 7.9 & 45.6 & 39.3 & 3.1 & 10.7 & 4.9 \\
\midrule

\multirow{6}{*}{Expression}
& \mbox{Nano Banana Pro}& \textbf{95.3} & \textbf{4.6} & \underline{0.5} & \underline{0.1} & \underline{0.5} & \textbf{1.4} & \textbf{9.4} \\
& \mbox{GPT-Image-1.5}   & 93.4 & 6.6 & \textbf{0.4} & \textbf{0.0} & \textbf{0.2} & 2.8 & 13.6 \\
& \mbox{Seedream 4.5}    & \underline{93.9} & \underline{6.0} & 0.6 & \underline{0.1} & \textbf{0.2} & \underline{2.3} & \underline{12.4} \\
& \mbox{HunyuanImage-3}  & 87.8 & 12.2 & 0.6 & \textbf{0.0} & \textbf{0.2} & \underline{2.3} & 25.5 \\
& \mbox{Qwen-Image-Edit} & 68.1 & 29.8 & 6.3 & 2.1 & 1.4 & 18.7 & 43.4 \\
& \mbox{OmniGen2}        & 66.9 & 32.9 & 3.0 & 0.2 & \underline{0.5} & 19.5 & 50.2 \\
\bottomrule
\end{tabular}
\end{table*}

Across all models, Table~\ref{tab:pattern_rates} reveals three regimes.
Nano Banana Pro and GPT-Image-1.5 are the most stable: they maintain the highest success rates and the lowest structured-pattern rates in nearly every dimension.
Seedream 4.5 is mixing-heavy with low drift rates, most clearly on face, where blending reaches 53.7\% and dominance 14.5\% despite only 12.1\% inconsistency. This suggests that Seedream 4.5 often preserves facial information from the references, but binds it to the wrong subject.
Hunyuan-Image-3.0-Instruct shows the opposite profile: it is drift-heavy rather than mixing-heavy, with face inconsistency 56.3\% and drift 45.6\%, but much lower face blending than Seedream 4.5. Its lower face-blending rate should not be interpreted as better identity binding; it mainly reflects weak facial preservation.
Qwen-Image-Edit-2511 and OmniGen2 are unstable on both counts, combining low success with high confusion and the highest swap, dominance, and blending rates on appearance and expression.
Notably, while Seedream 4.5 and Hunyuan-Image-3.0-Instruct remain competitive in holistic similarity and aesthetic scores, our binding diagnostics reveal severe mixing or drift patterns. This underscores the value of our binding metrics as a complementary perspective on holistic evaluation.

\paragraph{Per-dimension trends.}
\textbf{Face.}
Face identity most clearly separates mixing from decay.
Seedream 4.5 is face-mixing dominated (53.7\% blending, 14.5\% dominance), whereas Hunyuan-Image-3.0-Instruct is face-drift dominated (45.6\% drift, only 15.6\% confusion).
Qwen-Image-Edit-2511 and OmniGen2 add a third failure mode---assignment-style errors---with face swap rates of 17.1\% and 18.2\%, respectively, together with substantial dominance.

\textbf{Appearance.}
Appearance makes permutation-like errors easiest to see.
Qwen-Image-Edit-2511 and OmniGen2 show high appearance swap rates (22.9\% and 18.9\%), suggesting that appearance cues are often preserved but attached to the wrong subject.
By contrast, Nano Banana Pro and GPT-Image-1.5 keep appearance success above 94\% and appearance blending below 6\%, indicating comparatively stable appearance binding once identity remains intact.

\textbf{Pose.}
Pose remains challenging for all models, likely because it is only partially specified by text and is sensitive to distance and cropping. Nano Banana Pro achieves the highest pose success rate and generally exhibits less drift and confusion than the other models.

\textbf{Expression.}
Expression shows that low drift does not guarantee clean binding.
For the stronger closed models, drift is near zero, yet blending remains non-trivial (9.4--13.6\%), suggesting mild cross-subject coupling of facial expression even when self-consistency is preserved.
Hunyuan-Image-3.0-Instruct exhibits the same pattern more strongly (25.5\% blending with essentially zero drift), and the weaker open models escalate it into explicit collapse: dominance reaches 18.7\% for Qwen-Image-Edit-2511 and 19.5\% for OmniGen2, while blending reaches 43.4\% and 50.2\%, respectively, and drift remains comparatively small.
Expression therefore appears less limited by identity decay than by shared or copied affect across subjects.
\begin{figure*}[!t]
\centering
\includegraphics[width=1.0\textwidth]{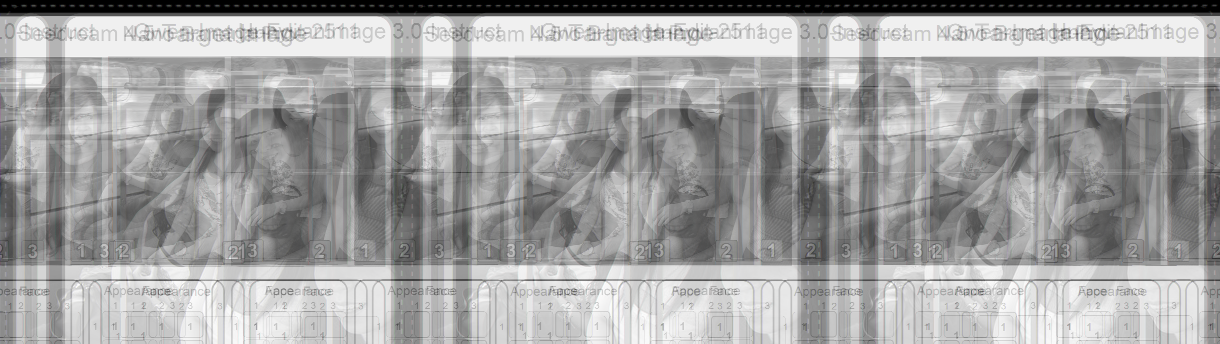}

\caption{Visualization of cross-subject attribute misbinding and our confusion-matrix diagnostics. Left: the ground-truth target image with slot-indexed subjects (1--3). Right: reconstructions from several multi-reference generators. For each output, we report the thresholded dimension-wise match matrices for \textit{Appearance} and \textit{Face}.}
\label{fig:diagnosis_cases}
\end{figure*}

\subsection{Qualitative and Quantitative Validation}

\paragraph{Qualitative Results.}
To make the diagnostic patterns concrete, Fig.~\ref{fig:diagnosis_cases} shows a representative example together with the thresholded match matrices for face and appearance.
In the face dimension, Nano Banana Pro retains strong diagonal matches for all three subjects, with only mild confusion between subjects 1 and 3. Seedream 4.5, by contrast, exhibits clear blending: the diagonal remains strong, but the off-diagonal face matches are also high, indicating averaged facial traits across subjects. Hunyuan-Image-3.0-Instruct and Qwen-Image-Edit-2511 illustrate identity drift on different subjects, where the generated faces no longer resemble their corresponding references.
In appearance, Nano Banana Pro, Seedream 4.5, and Hunyuan-Image-3.0-Instruct largely preserve the correct slot assignments, whereas Qwen-Image-Edit-2511 shows a clear leakage pattern for subject~2, combining the hairstyle of subject~2 with the clothing of subject~3. This case qualitatively matches the trends captured by our metrics. Additional cases are provided in \suppCaseRef.

\paragraph{Quantitative Verification with Human Judgments.}
To assess the reliability of the proposed diagnostics, we perform a meta-evaluation against human annotations. Following Sec.~\ref{sec:eval_patterns}, we compute the area under the curve (AUC) for diagonal scores $\Delta^{(d)}_{i,i}$ against consistency labels and for off-diagonal scores $\Delta^{(d)}_{i,j}$ against confusion labels, aggregated over all matched subjects. Across all four dimensions, our specialist-based metrics achieve higher AUC than VLM-as-a-judge baselines, indicating better agreement with human judgments. Detailed AUC values, the annotation protocol, and the VLM prompts are provided in the supplementary material.

\newcommand{\MultiBindContinuousResultsAppendix}{
\subsection{Dimension-wise continuous diagnostics}
\label{app:continuous_results}
\label{sec:results_dim}

Table~\ref{tab:dim_results} reports the threshold-free continuous summaries defined in Appendix~\ref{app:continuous_metrics}.
These metrics serve two purposes: they quantify the magnitude of diagonal degradation and off-diagonal mixing before thresholding, and they verify that the qualitative conclusions of the main text are not artifacts of the calibrated binarization.

\begin{table*}[t]
\centering
\small
\caption{Dimension-wise binding diagnostics (Appendix).
\textbf{Sim} is the mean diagonal of the raw gen-to-GT similarity matrix $S_{\mathrm{gen}}^{(d)}[i,i]$ (higher is better; not comparable across dimensions).
$D_{\mathrm{self}}^{(d)}$, $C_{\mathrm{worst}}^{(d)}$, $C_{\mathrm{mean}}^{(d)}$ (clamped), and $\mathrm{JS}^{(d)}$ are computed on baseline-corrected matrices (lower is better).}
\label{tab:dim_results}
\begin{tabular}{llccccc}
\toprule
Dimension $d$ & Model & Sim$\uparrow$ & $D_{\mathrm{self}}^{(d)}\downarrow$ & $C_{\mathrm{worst}}^{(d)}\downarrow$ & $C_{\mathrm{mean}}^{(d)}\downarrow$ & $\mathrm{JS}^{(d)}\downarrow$ \\
\midrule
Face identity & Nano Banana Pro & \textbf{0.3371} & \textbf{0.6629} & \underline{0.0449} & \underline{0.0266} & \textbf{0.0130} \\
& GPT-Image-1.5 & 0.2911 & 0.7089 & \textbf{0.0418} & \textbf{0.0244} & \underline{0.0145} \\
& Seedream 4.5 & \underline{0.2954} & \underline{0.7046} & 0.0835 & 0.0533 & 0.0165 \\
& Hunyuan-Image-3.0-Instruct & 0.0844 & 0.9156 & 0.0466 & 0.0281 & 0.0229 \\
& Qwen-Image-Edit-2511 & 0.1399 & 0.8601 & 0.0943 & 0.0578 & 0.0234 \\
& OmniGen2 & 0.1272 & 0.8728 & 0.0895 & 0.0567 & 0.0240 \\
\midrule
Appearance & Nano Banana Pro & \underline{0.8226} & \underline{0.1774} & \textbf{0.0168} & \textbf{0.0116} & \textbf{0.0011} \\
& GPT-Image-1.5 & \textbf{0.8282} & \textbf{0.1718} & \underline{0.0209} & \underline{0.0139} & \underline{0.0012} \\
& Seedream 4.5 & 0.8088 & 0.1912 & 0.0231 & 0.0144 & 0.0015 \\
& Hunyuan-Image-3.0-Instruct & 0.7598 & 0.2402 & 0.0491 & 0.0289 & 0.0025 \\
& Qwen-Image-Edit-2511 & 0.6564 & 0.3436 & 0.1140 & 0.0723 & 0.0069 \\
& OmniGen2 & 0.6720 & 0.3280 & 0.1162 & 0.0740 & 0.0068 \\
\midrule
Pose & Nano Banana Pro & \underline{0.5703} & \underline{0.4297} & \underline{0.0683} & 0.0522 & \textbf{0.0151} \\
& GPT-Image-1.5 & 0.5681 & 0.4319 & 0.0690 & \underline{0.0496} & \underline{0.0180} \\
& Seedream 4.5 & \textbf{0.5742} & \textbf{0.4258} & 0.0733 & 0.0554 & 0.0226 \\
& Hunyuan-Image-3.0-Instruct & 0.5504 & 0.4496 & 0.0715 & 0.0548 & 0.0254 \\
& Qwen-Image-Edit-2511 & 0.4538 & 0.5462 & \textbf{0.0641} & \textbf{0.0480} & 0.0313 \\
& OmniGen2 & 0.5126 & 0.4874 & 0.1136 & 0.0805 & 0.0333 \\
\midrule
Expression & Nano Banana Pro & \underline{0.8448} & \underline{0.1552} & \textbf{0.0119} & \textbf{0.0078} & \textbf{0.0006} \\
& GPT-Image-1.5 & \textbf{0.8536} & \textbf{0.1464} & 0.0163 & 0.0106 & \underline{0.0007} \\
& Seedream 4.5 & 0.8372 & 0.1628 & \underline{0.0150} & \underline{0.0095} & 0.0008 \\
& Hunyuan-Image-3.0-Instruct & 0.8425 & 0.1575 & 0.0251 & 0.0155 & 0.0009 \\
& Qwen-Image-Edit-2511 & 0.7537 & 0.2463 & 0.0518 & 0.0334 & 0.0024 \\
& OmniGen2 & 0.7817 & 0.2183 & 0.0658 & 0.0425 & 0.0025 \\
\bottomrule
\end{tabular}
\end{table*}

Across dimensions, the appendix view reinforces the same qualitative picture as the main-text pattern analysis.
Nano Banana Pro and GPT-Image-1.5 generally show the smallest diagonal degradation and the lowest JS, while the three open-source models are substantially weaker, especially on face identity and appearance.
Two details are worth highlighting.
First, pose is hard for all models: diagonal degradation is large across the board, but off-diagonal confusion remains comparatively moderate, matching the drift-heavy pose behavior seen in Table~\ref{tab:pattern_rates}.
Second, Hunyuan-Image-3.0-Instruct looks less confused than some weaker models on face off-diagonals, yet its face diagonal degradation is severe, which is exactly the drift-heavy regime highlighted in the main text.
Conversely, Seedream 4.5 stays relatively competitive on several diagonals but has noticeably larger face mixing, foreshadowing its blending-heavy pattern profile.
A small but instructive exception is pose for Qwen-Image-Edit-2511: it achieves the lowest continuous off-diagonal confusion on pose, yet its pose diagonal degradation and JS remain poor, indicating that its pose failures are dominated by self inconsistency rather than cross-subject transfer.
}

%% file: sections/conclusion.tex
\section{Conclusion}
We propose \textsc{MultiBind}, a benchmark for evaluating multi-reference, multi-subject image generation under complex instructions. By establishing deterministic slot correspondence via comprehensive annotations (e.g., masks, crops, references) and structured prompts, \textsc{MultiBind} enables the precise assessment of both reference preservation and edit accuracy. Furthermore, we introduce a specialist-based, dimension-wise confusion evaluation protocol that disentangles self-degradation from cross-subject interference. This exposes interpretable failure modes such as drift, swap, dominance, and blending. Evaluations across modern generators demonstrate that our metrics align with human judgment and reveal critical subject-attribute binding failures often obscured by holistic scores.

%% file: sections/supplementary.tex
\input{supplements/dataset_construction}

\section{Evaluation Details and Additional Metrics}

\subsection{Per-Model Generation Hyperparameter Settings}
\label{sec:gen_protocol}
\input{supplements/generation_protocol}

\subsection{Instance Extraction and Slot Matching}
\label{sec:eval_matching}
\input{supplements/matching_algorithm}

\subsection{Continuous Metrics}
\label{sec:continous_metrics}
\input{supplements/continuous_metrics}

\section{Aligning with Human Judgments}
\label{sec:align_human}
\input{supplements/annotations_auc}

\section{Ablation on the Reference-Image Generator}
\label{sec:supp_ablation_reference_generator}
\input{supplements/ablation}

\section{More Cases}
\label{sec:more_cases}
\input{supplements/more_cases}

%% file: supplements/dataset_construction.tex
\section{Dataset Construction Details}

\subsection{Implementation Details for the Image Pipeline}
\label{app:impl_details}

This section provides concrete implementation details for the image-side construction of \textsc{MultiBind}.

\paragraph{Source selection and filtering}
We curate multi-person scenes from four source pools: CIHP~\cite{gong2018pgn}, LV-MHP-v2~\cite{li2017mhp}, Objects365~\cite{shao2019objects365}, and COCO~\cite{lin2014coco}. We keep only instances with $N \in \{2,3,4\}$ human subjects.
When the source provides reliable per-person parsing or instance annotations, we use them directly.
Otherwise, we obtain instance masks and bounding boxes with a SAM-family segmenter~\cite{kirillov2023segmentanything}.

To suppress tiny or unreliable person instances, we apply explicit scale thresholds:
\begin{itemize}
    \item \textbf{Minimum area ratio:} $0.02$.
    \item \textbf{Minimum foreground size:} $500$ pixels.
\end{itemize}

For each retained subject, we compute a tight bounding box, export an alpha-masked crop, and construct the union foreground mask over all subjects in the image.
To support deterministic entity indexing in prompts and evaluation, subject slots are assigned from left to right according to the horizontal coordinates of mask centroids.

\paragraph{Canonicalization model, quality control (QC), and retry policy}
\label{app:canon_prompt}

We canonicalize each subject to a standardized full-body standing pose and a canonical facial expression while preserving identity and appearance.
Canonicalization is implemented as a generative transformation with strict VLM-based quality control.

\begin{itemize}
    \item \textbf{Transformation model:} Gemini 3 Pro Image (Nano Banana Pro)~\cite{google2025gemini3proimage_modelcard,google2026gemini3proimagepreview_docs}.
    \item \textbf{QC / verifier model:} Gemini 3 Pro ~\cite{google2025gemini3procard,google2026gemini3propreview_docs}.
    \item \textbf{QC acceptance:} a candidate is accepted only if \texttt{pass=true}, \\ \texttt{must\_regenerate=false}, and \texttt{score} $\ge 95$.
    \item \textbf{Max attempts:} up to 50 QC-driven regeneration attempts per subject.
\end{itemize}

When a subject is interacting with a prop, the prop is retained but repositioned so that it does not occlude the torso, waist, or legs.
Candidates that pass automatic QC are then screened by human annotators, and only those passing human review are retained.
If no acceptable candidate is obtained after the retry budget is exhausted, the subject is discarded.

\paragraph{Background inpainting model and QC}

We derive the background reference by removing all subjects specified by the union mask and inpainting the masked region.
The inpainting process requires complete removal of the masked subjects and all associated visual traces, including cast shadows, reflections, and other person-induced artifacts, while preserving all unrelated background content and maintaining globally consistent lighting and perspective.

\begin{itemize}
    \item \textbf{Inpainting model:} Gemini 3 Pro Image (Nano Banana Pro)~\cite{google2025gemini3proimage_modelcard,google2026gemini3proimagepreview_docs}.
    \item \textbf{QC model:} Gemini 3 Pro ~\cite{google2025gemini3procard,google2026gemini3propreview_docs}.
    \item \textbf{Max attempts:} up to 5 candidates per image; the first candidate that passes QC is kept.
\end{itemize}

As with subject canonicalization, candidates that pass automatic QC are manually checked before release.
If no candidate passes review, the instance is discarded.
The background reference is generated at an aspect-ratio-preserving resolution.

\subsection{Captioning, Verification, and Prompt Compilation}
\label{sec:supp_caption_prompt}
\input{supplements/caption_prompt}

\subsection{Dataset Statistics}
\label{app:dataset_stats}

Prompt length is measured in \textbf{words} using regex tokenization
(\texttt{\textbackslash b\textbackslash w+\textbackslash b}) on the standardized prompt text.
Overall, \textsc{MultiBind} contains 508 instances and 1,527 human subjects.
Table~\ref{tab:multibind_person_count} summarizes the distribution of the number of persons per instance,
Table~\ref{tab:multibind_source_distribution} reports the source composition,
Table~\ref{tab:multibind_prompt_length} presents prompt-length statistics in words,
and Table~\ref{tab:multibind_resolution} reports image-resolution statistics for the long and short sides in pixels.

\begin{table}[!t]
    \centering
    \caption{Distribution of person counts in \textsc{MultiBind}.}
    \label{tab:multibind_person_count}
    \small
    \begin{tabular}{lrr}
        \toprule
        $N$ & \#instances & ratio \\
        \midrule
        2 & 118 & 0.232 \\
        3 & 269 & 0.530 \\
        4 & 121 & 0.238 \\
        \bottomrule
    \end{tabular}
\end{table}

\begin{table}[!t]
    \centering
    \caption{Source distribution of instances in \textsc{MultiBind}.}
    \label{tab:multibind_source_distribution}
    \small
    \begin{tabular}{lrr}
        \toprule
        Source & \#instances & ratio \\
        \midrule
        CIHP & 204 & 0.402 \\
        LV-MHP-v2 & 154 & 0.303 \\
        Objects365 & 133 & 0.262 \\
        COCO & 17 & 0.033 \\
        \bottomrule
    \end{tabular}
\end{table}

\begin{table}[!t]
    \centering
    \caption{Prompt length statistics of \textsc{MultiBind}, measured in words.}
    \label{tab:multibind_prompt_length}
    \small
    \begin{tabular}{lr}
        \toprule
        Metric & Value \\
        \midrule
        Count & 508 \\
        Mean & 473.84 \\
        Median & 473.50 \\
        $p10$ & 350.70 \\
        $p90$ & 592.00 \\
        Min & 255 \\
        Max & 761 \\
        \bottomrule
    \end{tabular}
\end{table}

\begin{table}[!t]
    \centering
    \caption{Resolution statistics of \textsc{MultiBind}. Long-side and short-side values are reported in pixels.}
    \label{tab:multibind_resolution}
    \small
    \begin{tabular}{llr}
        \toprule
        Side & Metric & Value \\
        \midrule
        Long & Count & 508 \\
        Long & Mean & 624.38 \\
        Long & Median & 500.00 \\
        Long & $p10$ & 500.00 \\
        Long & $p90$ & 805.10 \\
        Long & Min & 256 \\
        Long & Max & 4272 \\
        \midrule
        Short & Count & 508 \\
        Short & Mean & 439.07 \\
        Short & Median & 375.00 \\
        Short & $p10$ & 331.70 \\
        Short & $p90$ & 577.00 \\
        Short & Min & 188 \\
        Short & Max & 2848 \\
        \bottomrule
    \end{tabular}
\end{table}

%% file: supplements/caption_prompt.tex
This supplementary section documents the full pipeline used to construct structured captions and final natural-language prompts for evaluation.
The exact prompts used in the three LLM stages are provided in
\texttt{caption.txt},
\texttt{evaluation.txt}, and
\texttt{review.txt}.

\paragraph{Overview.}
For each instance, we use a three-stage
\emph{caption--evaluation--review} pipeline.
A caption model first produces a dense structured caption from the target image.
An evaluation model then checks the caption against the same image and proposes minimal edits only for potentially inconsistent fields.
A review model finally adjudicates each proposed edit with one of three decisions:
\texttt{accept}, \texttt{reject}, or \texttt{human\_review}.
The finalized structured caption is then compiled into the natural-language prompt using a deterministic, rule-based compiler rather than free-form LLM rewriting.

\paragraph{Structured captioning.}
Grounding \textsc{MultiBind} in real target images enables long, attribute-rich, entity-indexed prompts while preserving global semantic coherence.
Because our task is \emph{reconstruction} and evaluation is performed against $I_{\mathrm{gt}}$, caption errors can
(i) make an instance ill-posed by requesting attributes absent from $I_{\mathrm{gt}}$ and
(ii) introduce noise into controllability analysis.
We therefore annotate each instance with a dense, structured caption schema that covers:
(i) global scene cues, including layout, lighting, and style;
(ii) per-subject attributes, including both $s^{\text{edit}}$ and salient $s^{\text{preserve}}$ factors; and
(iii) inter-subject relations, including relative position, interaction and physical contact, and occlusion ordering.
Captions are generated and verified using Gemini~3~Pro.

To reduce ambiguity, we adopt explicit conventions for left and right, viewpoint, and relation naming.
In particular, relative-position descriptors are computed deterministically from ground-truth geometry and then verbalized under a fixed viewpoint convention.

\paragraph{Evaluation.}
The second-stage evaluation inspects the structured fields produced in the first stage, identifies only fields that appear inconsistent with $I_{\mathrm{gt}}$, and proposes minimal replacements for those fields.
This restricted-edit protocol reduces cascading changes, preserves the image-grounded structure of the original caption, and makes later verification more targeted and auditable.

\paragraph{Review.}
For each flagged field, the review prompt receives the original value, the proposed fix, and the target image, and outputs one of three decisions.
\texttt{accept} means the proposed fix is adopted,
\texttt{reject} means the system reverts to the original value, and
\texttt{human\_review} reserves the field for manual inspection.
The review stage is intentionally local: it verifies only flagged edits instead of re-captioning the entire instance.
This makes the adjudication process easier to audit and reduces additional hallucinations during correction.

\paragraph{Blind human audit of review decisions.}
We additionally perform a blind human audit to estimate the quality of automatic adjudication.
The audit set contains 100 model-accepted edits and 100 model-rejected edits.
Human annotators confirm 92/100 accepted edits and 87/100 rejected edits.
All cases marked as \texttt{human\_review} are manually adjudicated before final release.

\paragraph{Prompt compilation.}
We compile the finalized structured captions into the natural-language prompt $p$ using a deterministic, rule-based compiler.
Each caption field is first normalized into a canonical textual form and then inserted into a fixed template.
Each subject is bound to a fixed subject slot (e.g., ``Subject~A'', ``Subject~B'', and ``Subject~C''), and all per-subject attributes and relations are rendered with explicit slot mentions.
We additionally enforce a minimal-change rule: attributes not explicitly requested in $p$ should be preserved from the corresponding subject reference $r_i^{\mathrm{subject}}$.

We prefer deterministic compilation over free-form LLM rewriting for reproducibility and control.
Unconstrained rewriting can introduce paraphrase drift, hallucinated details, or accidental attribute changes, which would confound evaluation.
Although template-based prompts may be less stylistically fluent, they are stable across instances and across runs.

\paragraph{Validation checks.}
Before releasing a compiled prompt, we apply lightweight validation checks.
These checks include:
(i) every subject slot is referenced in its dedicated prompt block;
(ii) relation statements reference valid subject indices and follow the fixed viewpoint convention; and
(iii) prompts with contradictory categorical fields, missing mandatory fields, or malformed formatting are rejected.
Prompt templates, compilation rules, and examples are provided in this supplementary section and in the accompanying code files.

\paragraph{Example.}
We include one full example consisting of
(i) the finalized structured caption and
(ii) the compiled natural-language prompt.
This example illustrates how dense caption fields are normalized, how geometric cues are converted into natural spatial language, and how all subject-specific and relational information is assembled into the final prompt.

\medskip
\noindent\textbf{Structured caption example.}

See \texttt{caption\_example.json}

\medskip
\noindent\textbf{Compiled prompt example.}

See \texttt{prompt\_example.txt}

%% file: supplements/generation_protocol.tex
The settings below correspond to the six generators evaluated in the main paper: Gemini 3 Pro Image (Nano Banana Pro)~\cite{google2025gemini3proimage_modelcard,google2026gemini3proimagepreview_docs}, GPT-Image-1.5~\cite{gpt}, Seedream 4.5~\cite{seedream}, HunyuanImage-3.0-Instruct~\cite{hunyuan}, Qwen-Image-Edit-2511~\cite{qwen}, and OmniGen2~\cite{omnigen2}.

\paragraph{Shared reconstruction setup.}
For each sample, the model receives $N\in\{2,3,4\}$ subject references, one background reference, and the prompt. Images are passed to the model in the order subject $1\ldots N$ followed by the background reference.

\begin{table*}[!t]
\centering
\small
\caption{Generation settings for Qwen-Image-Edit-2511.}
\label{tab:gen-settings-qwen}
\begin{tabular}{@{}p{0.23\textwidth}p{0.73\textwidth}@{}}
\toprule
Item & Setting \\
\midrule

Precision & \texttt{dtype=bfloat16} \\
Denoising steps & \texttt{num\_inference\_steps=40} \\
CFG settings & \texttt{true\_cfg\_scale=4.0}, \texttt{guidance\_scale=1.0} \\

\bottomrule
\end{tabular}
\end{table*}

\begin{table}[!t]
\centering
\small
\caption{Generation settings for OmniGen2.}
\label{tab:gen-settings-omnigen}
\begin{tabularx}{\columnwidth}{@{}p{0.28\columnwidth}X@{}}
\toprule
Item & Setting \\
\midrule

Precision & \texttt{dtype=bfloat16}\\
Denoising steps & \texttt{num\_inference\_steps=50} \\
Guidance settings & \texttt{text\_guidance\_scale=5.0}, \texttt{image\_guidance\_scale=2.0}\\

\bottomrule
\end{tabularx}
\end{table}

%% file: supplements/matching_algorithm.tex
Given a generated image $I_{\mathrm{gen}}$ and its ground-truth image $I_{\mathrm{gt}}$ containing $N$ annotated subjects, we need to match the generated subjects with the subjects in the original image in order to perform subsequent comparison and analysis. We first load the ground-truth instance masks
$\{M_i^{\mathrm{gt}}\}_{i=1}^{N}$ and compute their bounding boxes
$\{B_i^{\mathrm{gt}}\}_{i=1}^{N}$.
We then run a SAM-family segmenter~\cite{kirillov2023segmentanything} on $I_{\mathrm{gen}}$ with the text prompt \texttt{``person''} and a confidence threshold $\tau_{\mathrm{conf}}$, obtaining candidate detections
$\{(M_j^{\mathrm{det}}, B_j^{\mathrm{det}}, s_j)\}_{j=1}^{K}$,
where $M_j^{\mathrm{det}}$ is the predicted mask,
$B_j^{\mathrm{det}}$ is the corresponding bounding box, and
$s_j$ is the detection confidence.
To compare masks in a common coordinate system, the generated image is resized to the ground-truth resolution before segmentation and matching.

Our default matching rule is a deterministic \texttt{topk\_area\_ltr} procedure.
Let
\[
a_i=\frac{\mathrm{area}(B_i^{\mathrm{gt}})}{|I_{\mathrm{gt}}|}
\]
denote the normalized bounding-box area of ground-truth subject $i$.
We define an adaptive minimum detection-size threshold
\[
T_{\mathrm{area}}=\alpha \cdot \min_i a_i,
\]
where $\alpha=0.35$ in our implementation.
A detection $j$ is kept only if its normalized bounding-box area satisfies
\[
\frac{\mathrm{area}(B_j^{\mathrm{det}})}{|I_{\mathrm{gen}}|} \ge T_{\mathrm{area}}
\]
and its confidence satisfies $s_j \ge \tau_{\mathrm{conf}}$.

We further remove near-duplicate detections greedily.
Candidate detections are sorted by decreasing bounding-box area (with confidence as a tie-breaker), and a detection is discarded if its mask has large overlap with any already-kept detection.
Specifically, for two masks $M_a$ and $M_b$, we define
\[
\operatorname{ovl}(M_a,M_b)
=
\frac{|M_a \cap M_b|}{\min(|M_a|,|M_b|)},
\]
and treat two detections as duplicates when
$\operatorname{ovl}(M_a,M_b)\ge\tau_{\mathrm{dup}}$.
In our implementation, $\tau_{\mathrm{dup}}=0.5$.

If at least $N$ detections remain after filtering and de-duplication, we keep the top-$N$ detections by area and perform slot assignment using a left-to-right ordering heuristic.
More concretely, we sort both the ground-truth subjects and the selected detections by the $x$-coordinate of their mask centroids, breaking ties by centroid $y$ and then box $x_1$.
We then assign detections to subject slots by rank.
This gives a deterministic relative-position matching rule that is well aligned with the left-to-right subject layout used in our multi-person scenes.

If fewer than $N$ detections remain after filtering and de-duplication, we fall back to Hungarian matching on the mask-IoU matrix between the ground-truth subjects and the surviving detections.
The fallback is restricted to the filtered detections only and maximizes total IoU without imposing an additional IoU threshold, thereby recovering as many assignments as possible when detections are missing. The complete process is described in Algorithm \ref{alg:slot_matching}.

Let $\mathcal{M}\subseteq\{1,\ldots,N\}$ denote the subset of ground-truth slots that receive an assignment.
We report the assignment rate $|\mathcal{M}|/N$ and the mean mask IoU over assigned pairs as auxiliary localization diagnostics.
For binding-related metrics, different models may assign different subsets of slots.
We therefore evaluate each instance on the intersection of assigned slots across compared models, ensuring identical subject subsets.

\begin{algorithm}[!t]
\caption{Instance extraction and slot matching (\texttt{topk\_area\_ltr})}
\label{alg:slot_matching}
\begin{algorithmic}[1]
\Require Generated image $I_{\mathrm{gen}}$, ground-truth masks $\{M_i^{\mathrm{gt}}\}_{i=1}^{N}$, confidence threshold $\tau_{\mathrm{conf}}$, area factor $\alpha$, de-duplication threshold $\tau_{\mathrm{dup}}$
\Ensure Assignment set $\Pi$

\State Resize $I_{\mathrm{gen}}$ to the ground-truth resolution
\State Run the segmenter on $I_{\mathrm{gen}}$ with text prompt ``person''
\State Obtain detections $\mathcal{D}=\{(M_j^{\mathrm{det}}, B_j^{\mathrm{det}}, s_j)\}_{j=1}^{K}$
\State Remove detections with empty masks or $s_j < \tau_{\mathrm{conf}}$

\State Compute
\[
T_{\mathrm{area}} \gets
\alpha \cdot \min_i \frac{\mathrm{area}(B_i^{\mathrm{gt}})}{|I_{\mathrm{gt}}|}
\]

\State Keep candidates
\[
\mathcal{C} \gets
\left\{
 j :
 \frac{\mathrm{area}(B_j^{\mathrm{det}})}{|I_{\mathrm{gen}}|} \ge T_{\mathrm{area}}
\right\}
\]

\State Sort $\mathcal{C}$ by decreasing $\mathrm{area}(B_j^{\mathrm{det}})$, then by decreasing $s_j$
\State $\mathcal{K} \gets [\,]$

\For{$j \in \mathcal{C}$}
    \If{$\frac{|M_j^{\mathrm{det}}\cap M_k^{\mathrm{det}}|}{\min(|M_j^{\mathrm{det}}|,|M_k^{\mathrm{det}}|)} < \tau_{\mathrm{dup}}$ for all $k \in \mathcal{K}$}
        \State Append $j$ to $\mathcal{K}$
    \EndIf
\EndFor

\If{$|\mathcal{K}| \ge N$}
    \State $\mathcal{S} \gets$ first $N$ elements of $\mathcal{K}$ \Comment{top-$N$ by area}
    \State Sort GT indices $\mathcal{G}$ by centroid-$x$ of $M_i^{\mathrm{gt}}$
    \State Break ties in $\mathcal{G}$ by centroid-$y$, then by $x_1(B_i^{\mathrm{gt}})$
    \State Sort selected detections $\mathcal{S}$ by centroid-$x$ of $M_j^{\mathrm{det}}$
    \State Break ties in $\mathcal{S}$ by centroid-$y$, then by $x_1(B_j^{\mathrm{det}})$
    \State $\Pi \gets \{(\mathcal{G}[t], \mathcal{S}[t])\}_{t=1}^{N}$
\Else
    \State Compute IoU matrix $A_{ij}=\operatorname{IoU}(M_i^{\mathrm{gt}}, M_j^{\mathrm{det}})$ for $j\in\mathcal{K}$
    \State $\Pi \gets \textsc{Hungarian}(1-A)$ \Comment{maximize total IoU over surviving detections}
\EndIf

\State \Return $\Pi$
\end{algorithmic}
\end{algorithm}

%% file: supplements/continuous_metrics.tex
In the main paper (Sec.~4), we convert similarity deltas into calibrated binary indicators to obtain interpretable failure patterns such as drift, swap, dominance, and blending.
While these thresholded diagnostics are useful for discrete pattern analysis, it is also informative to examine the underlying \emph{continuous} similarity changes before binarization.

In this section we therefore report continuous confusion summaries derived directly from the delta matrices.
These metrics quantify how much similarity structure changes under generation, separating
(i) generic self-degradation from
(ii) cross-subject feature mixing.

Recall from Sec.~4.1 that for each attribute dimension $d$ we compute the baseline-corrected similarity matrix
\begin{equation}
\Delta^{(d)} = S^{(d)}_{\mathrm{gen}} - S^{(d)}_{\mathrm{gt}},
\end{equation}
where $S^{(d)}_{\mathrm{gen}}$ measures similarity between generated subjects and ground-truth subjects,
and $S^{(d)}_{\mathrm{gt}}$ captures the intrinsic similarity among ground-truth subjects themselves.
The subtraction isolates changes introduced by generation and removes biases caused by naturally similar subjects.

The diagonal entries $\Delta^{(d)}[i,i]$ measure how much the generated subject in slot $i$ retains its own attributes,
while off-diagonal entries $\Delta^{(d)}[i,j]$ ($j\neq i$) indicate whether the generated subject moves toward another ground-truth subject beyond the baseline similarity.

From $\Delta^{(d)}$ we derive several scalar summaries.

\paragraph{Self-degradation.}

We measure the average diagonal drop as

\begin{equation}
D_{\mathrm{self}}^{(d)}
=
-\frac{1}{|\mathcal{I}^{(d)}|}
\sum_{i\in\mathcal{I}^{(d)}} \Delta^{(d)}[i,i].
\end{equation}

This quantity captures generic quality degradation, i.e., how much a generated subject loses similarity to its own ground-truth counterpart, without attributing the error to any particular competing subject.

\paragraph{Mean cross-subject mixing.}

Diffuse feature leakage across multiple subjects is summarized by

\begin{equation}
C_{\mathrm{mean}}^{(d)}
=
\frac{1}{|\mathcal{I}^{(d)}|}
\sum_{i\in\mathcal{I}^{(d)}}
\frac{1}{|\mathcal{V}^{(d)}|-1}
\sum_{j\in\mathcal{V}^{(d)},\, j\neq i}
\big[\Delta^{(d)}[i,j]\big]_+ ,
\end{equation}

where $[x]_+ = \max(x,0)$ ensures that decreases in similarity do not cancel increases.
This metric captures \emph{diffuse mixing}, where attributes bleed across several subjects.

\paragraph{Worst confusion.}

To capture strong pairwise interference, we also report

\begin{equation}
C_{\mathrm{worst}}^{(d)}
=
\frac{1}{|\mathcal{I}^{(d)}|}
\sum_{i\in\mathcal{I}^{(d)}}
\max_{j\in\mathcal{V}^{(d)},\, j\neq i}
\big[\Delta^{(d)}[i,j]\big]_+ .
\end{equation}

This metric highlights near-swaps or strong one-to-one confusion between specific subject pairs.

Finally, to summarize the overall redistribution of similarity mass across candidate subjects,
we also report the row-wise Jensen--Shannon shift $\mathrm{JS}^{(d)}$ defined in Sec.~4.3 of the main paper.

\subsubsection{Dimension-wise binding results}

Table~\ref{tab:dim_results} reports these continuous metrics for all four attribute dimensions and the six evaluated generators: Nano Banana Pro~\cite{google2025gemini3proimage_modelcard,google2026gemini3proimagepreview_docs}, GPT-Image-1.5~\cite{gpt}, Seedream 4.5~\cite{seedream}, HunyuanImage-3.0-Instruct~\cite{hunyuan}, Qwen-Image-Edit-2511~\cite{qwen}, and OmniGen2~\cite{omnigen2}.

\begin{table*}[t]
\centering
\small
\caption{Dimension-wise binding diagnostics.
\textbf{Sim} is the mean diagonal of the raw gen-to-GT similarity matrix $S_{\mathrm{gen}}^{(d)}[i,i]$ (higher is better; not comparable across dimensions).
$D_{\mathrm{self}}^{(d)}$, $C_{\mathrm{worst}}^{(d)}$, $C_{\mathrm{mean}}^{(d)}$ (clamped), and $\mathrm{JS}^{(d)}$ are computed on baseline-corrected matrices (lower is better).
HunyuanImage-3 denotes HunyuanImage-3.0-Instruct; Qwen-Image-Edit denotes Qwen-Image-Edit-2511.}
\label{tab:dim_results}
\begin{tabular}{llccccc}
\toprule
Dimension $d$ & Model & Sim$\uparrow$ & $D_{\mathrm{self}}^{(d)}\downarrow$ & $C_{\mathrm{worst}}^{(d)}\downarrow$ & $C_{\mathrm{mean}}^{(d)}\downarrow$ & $\mathrm{JS}^{(d)}\downarrow$ \\
\midrule
Face identity & \mbox{Nano Banana Pro} & \textbf{0.3371} & \textbf{0.6629} & \underline{0.0449} & \underline{0.0266} & \textbf{0.0130} \\
& \mbox{GPT-Image-1.5} & 0.2911 & 0.7089 & \textbf{0.0418} & \textbf{0.0244} & \underline{0.0145} \\
& \mbox{Seedream 4.5} & \underline{0.2954} & \underline{0.7046} & 0.0835 & 0.0533 & 0.0165 \\
& \mbox{HunyuanImage-3} & 0.0844 & 0.9156 & 0.0466 & 0.0281 & 0.0229 \\
& \mbox{Qwen-Image-Edit} & 0.1399 & 0.8601 & 0.0943 & 0.0578 & 0.0234 \\
& \mbox{OmniGen2} & 0.1272 & 0.8728 & 0.0895 & 0.0567 & 0.0240 \\
\midrule
Appearance & \mbox{Nano Banana Pro} & \underline{0.8226} & \underline{0.1774} & \textbf{0.0168} & \textbf{0.0116} & \textbf{0.0011} \\
& \mbox{GPT-Image-1.5} & \textbf{0.8282} & \textbf{0.1718} & \underline{0.0209} & \underline{0.0139} & \underline{0.0012} \\
& \mbox{Seedream 4.5} & 0.8088 & 0.1912 & 0.0231 & 0.0144 & 0.0015 \\
& \mbox{HunyuanImage-3} & 0.7598 & 0.2402 & 0.0491 & 0.0289 & 0.0025 \\
& \mbox{Qwen-Image-Edit} & 0.6564 & 0.3436 & 0.1140 & 0.0723 & 0.0069 \\
& \mbox{OmniGen2} & 0.6720 & 0.3280 & 0.1162 & 0.0740 & 0.0068 \\
\midrule
Pose & \mbox{Nano Banana Pro} & \underline{0.5703} & \underline{0.4297} & \underline{0.0683} & 0.0522 & \textbf{0.0151} \\
& \mbox{GPT-Image-1.5} & 0.5681 & 0.4319 & 0.0690 & \underline{0.0496} & \underline{0.0180} \\
& \mbox{Seedream 4.5} & \textbf{0.5742} & \textbf{0.4258} & 0.0733 & 0.0554 & 0.0226 \\
& \mbox{HunyuanImage-3} & 0.5504 & 0.4496 & 0.0715 & 0.0548 & 0.0254 \\
& \mbox{Qwen-Image-Edit} & 0.4538 & 0.5462 & \textbf{0.0641} & \textbf{0.0480} & 0.0313 \\
& \mbox{OmniGen2} & 0.5126 & 0.4874 & 0.1136 & 0.0805 & 0.0333 \\
\midrule
Expression & \mbox{Nano Banana Pro} & \underline{0.8448} & \underline{0.1552} & \textbf{0.0119} & \textbf{0.0078} & \textbf{0.0006} \\
& \mbox{GPT-Image-1.5} & \textbf{0.8536} & \textbf{0.1464} & 0.0163 & 0.0106 & \underline{0.0007} \\
& \mbox{Seedream 4.5} & 0.8372 & 0.1628 & \underline{0.0150} & \underline{0.0095} & 0.0008 \\
& \mbox{HunyuanImage-3} & 0.8425 & 0.1575 & 0.0251 & 0.0155 & 0.0009 \\
& \mbox{Qwen-Image-Edit} & 0.7537 & 0.2463 & 0.0518 & 0.0334 & 0.0024 \\
& \mbox{OmniGen2} & 0.7817 & 0.2183 & 0.0658 & 0.0425 & 0.0025 \\
\bottomrule
\end{tabular}
\end{table*}

%% file: supplements/annotations_auc.tex
We complement the automatic evaluation in the main paper with a human-labeled validation subset.
The goal of this section is twofold:
(i) to calibrate the per-dimension thresholds used in Sec.~4.2 of the main paper, and
(ii) to verify whether our pair-level delta scores align with human judgments better than strong VLM judges.

\subsection{Human Annotations}
\label{sec:human_annotations}

We manually annotate a sampled subset of $3{,}664$ subject--subject pairs drawn from the evaluated generations, covering all six generators and all four dimensions.
This includes $1{,}132$ diagonal self-consistency pairs and $2{,}532$ off-diagonal confusion pairs.
Equivalently, each dimension contributes 283 self-consistency pairs and 633 confusion pairs.

We collect two types of binary human labels:
\begin{itemize}
    \item \textbf{Self-consistency} (diagonal): for each queried diagonal pair $(i,i)$, whether the generated subject assigned to slot $i$ is consistent with its own ground-truth subject in dimension $d$.
    \item \textbf{Cross-subject confusion} (off-diagonal): for each queried off-diagonal pair $(i,j)$ with $j \neq i$, whether the generated subject assigned to slot $i$ appears confused with the wrong ground-truth subject $j$ in dimension $d$.
\end{itemize}

All labels follow a fixed written protocol with dimension-specific criteria.

\subsection{Per-Dimension Threshold Calibration}
\label{sec:human_thresholds}

Following Sec.~4.2 of the main paper, we calibrate one diagonal self-consistency threshold and one off-diagonal confusion threshold for each dimension using the labeled subset above.
Table~\ref{tab:thresholds} reports the thresholds used to binarize the delta matrices in the main paper.

\begin{table}[t]
\centering
\small
\caption{Per-dimension thresholds calibrated on the human-labeled subset.
$\tau^{(d)}_{\mathrm{cons}}$ is used for diagonal self-consistency and $\tau^{(d)}_{\mathrm{conf}}$ for off-diagonal confusion.}
\label{tab:thresholds}
\begin{tabular}{lcc}
\toprule
Dimension $d$ & $\tau^{(d)}_{\mathrm{cons}}$ & $\tau^{(d)}_{\mathrm{conf}}$ \\
\midrule
Face identity & $-0.9111$ & $0.1086$ \\
Appearance    & $-0.3662$ & $0.1117$ \\
Pose          & $-0.5289$ & $0.2912$ \\
Expression    & $-0.4203$ & $0.0714$ \\
\bottomrule
\end{tabular}
\end{table}

\subsection{VLM Annotation Procedure}
\label{sec:vlm_annotation}

As judge baselines, we evaluate two strong VLMs: Gemini 2.5 Pro~\cite{google2026gemini25pro_docs} and GPT-5.2~\cite{openai2026gpt5model}.
For each human-labeled query, we construct the analogous VLM question at the same granularity:
(a) a \emph{self-consistency} question for a diagonal pair $(i,i)$, and
(b) a \emph{cross-subject confusion} question for an off-diagonal pair $(i,j)$ with $j\neq i$.
Each prompt specifies the target dimension $d$ and asks the VLM to return a scalar judgment score for the queried relation.
We use the returned score directly for AUC computation against the same human labels described above.
The full prompt templates are provided in \texttt{vlm\_annotate\_confusion.txt}.

\subsection{AUC Against Human Labels}
\label{sec:auc_against_humans}

Using the pair-level human labels, we measure how well automatic scores rank positive pairs above negative pairs.
For our method, we use the raw pair-level delta scores themselves:
$\Delta^{(d)}[i,i]$ for self-consistency and $\Delta^{(d)}[i,j]$ for cross-subject confusion.
For the VLM baselines, we use the judge scores from Sec.~\ref{sec:vlm_annotation}.
We report ROC-AUC separately for each dimension and for the two tasks.

\begin{table*}[t]
\centering
\small
\caption{AUC validation against human labels using single-pass VLM judgments (higher is better).
Left: self-consistency (diagonal pairs). Right: cross-subject confusion (off-diagonal pairs).
We compare our pair-level delta scores with two VLM judges, Gemini 2.5 Pro and GPT-5.2.}
\label{tab:auc_validation_humans}
\begin{tabular}{lccc|ccc}
\toprule
& \multicolumn{3}{c|}{Self-consistency AUC$\uparrow$} & \multicolumn{3}{c}{Cross-subject confusion AUC$\uparrow$} \\
\cmidrule(lr){2-4}\cmidrule(lr){5-7}
Dimension $d$ & Ours ($\Delta_{i,i}$) & Gemini 2.5 Pro & GPT-5.2 & Ours ($\Delta_{i,j}$) & Gemini 2.5 Pro & GPT-5.2\\
\midrule
Face identity & 0.8695 & 0.7823 & 0.7376 & 0.8455 & 0.6894 & 0.6076\\
Appearance    & 0.9292 & 0.8069 & 0.8184 & 0.8979 & 0.7586 & 0.7163\\
Pose          & 0.7608 & 0.6505 & 0.6192 & 0.7112 & 0.5941 & 0.4768\\
Expression    & 0.8034 & 0.8393 & 0.7107 & 0.7356 & 0.6705 & 0.6318\\
\bottomrule
\end{tabular}
\end{table*}

Table~\ref{tab:auc_validation_humans} reports the final AUC comparison between our specialist-based scores and the VLM judge baselines for both tasks.

%% file: supplements/ablation.tex
Most \textsc{MultiBind} instances use synthetic reference images generated by Nano Banana Pro during dataset construction, while Nano Banana Pro is itself one of the models evaluated in the benchmark. This raises a potential concern: the benchmark might unfairly favor a model when the synthetic references are produced by the same model family. To test this, we build an ablation subset and regenerate the reference images with three different generators: Nano Banana Pro, GPT-Image-1.5, and Seedream 4.5~\cite{google2025gemini3proimage_modelcard,google2026gemini3proimagepreview_docs,gpt,seedream}. We then rerun the same evaluation pipeline on the same subset. The target images, prompts, matching protocol, specialists, and binarization thresholds are kept fixed; only the reference-image generator changes.

Because this ablation uses a smaller subset and the reference synthesis itself is stochastic, we do not over-interpret small numerical differences. Instead, we focus on whether the qualitative conclusions and failure patterns remain stable across the three reference-generation choices. If the main-paper results were dominated by a same-model bias in reference synthesis, we would expect each model to improve specifically and consistently when evaluated with references produced by itself.

\subsection{Holistic Results}
\label{sec:supp_ablation_reference_generator_holistic}

Tables~\ref{tab:ablation_holistic_banana_ref}--\ref{tab:ablation_holistic_seedream_ref} report holistic metrics on the same ablation subset under three reference-image generators. 

\begin{table*}[t]
\centering
\small
\caption{Holistic metrics on the ablation subset of \textsc{MultiBind} when subject reference images are synthesized by Nano Banana Pro. The ``Matched'' metric reports the number of aligned subject slots within the all-model success intersection for this reference-generator setting.}
\label{tab:ablation_holistic_banana_ref}
\begin{tabular}{lccccccc}
\toprule
Model & FID$\downarrow$ & CLIP-I$\uparrow$ & DINO$\uparrow$ & AES$\uparrow$ & JS$\downarrow$ & Matched$\uparrow$ & Mean IoU$\uparrow$\\
\midrule
Nano Banana Pro & \textbf{135.67} & \textbf{0.88} & \textbf{0.83} & 5.00 & \textbf{0.0076} & \underline{254} & \underline{0.42} \\
GPT-Image-1.5 & \underline{139.38} & 0.86 & 0.78 & \textbf{5.46} & \underline{0.0076} & \textbf{255} & 0.42 \\
Seedream 4.5 & 140.84 & \underline{0.87} & \underline{0.78} & 4.86 & 0.0082 & 252 & 0.33 \\
HunyuanImage-3.0-Instruct & 143.30 & 0.81 & 0.76 & \underline{5.01} & 0.0109 & \textbf{255} & \textbf{0.43} \\
Qwen-Image-Edit-2511 & 169.20 & 0.78 & 0.59 & 4.60 & 0.0142 & 218 & 0.28 \\
OmniGen2 & 156.82 & 0.80 & 0.69 & 4.54 & 0.0152 & 218 & 0.29 \\
\bottomrule
\end{tabular}
\end{table*}

\begin{table*}[t]
\centering
\small
\caption{Holistic metrics on the ablation subset of \textsc{MultiBind} when subject reference images are synthesized by GPT-Image-1.5. The ``Matched'' metric reports the number of aligned subject slots within the all-model success intersection for this reference-generator setting.}
\label{tab:ablation_holistic_gpt_ref}
\begin{tabular}{lccccccc}
\toprule
Model & FID$\downarrow$ & CLIP-I$\uparrow$ & DINO$\uparrow$ & AES$\uparrow$ & JS$\downarrow$ & Matched$\uparrow$ & Mean IoU$\uparrow$\\
\midrule
Nano Banana Pro & 153.37 & \underline{0.85} & \textbf{0.78} & \underline{5.49} & \underline{0.0079} & \underline{253} & 0.37 \\
GPT-Image-1.5 & \underline{145.48} & 0.84 & \underline{0.76} & \textbf{5.58} & \textbf{0.0077} & \underline{253} & \underline{0.40} \\
Seedream 4.5 & \textbf{143.68} & \textbf{0.85} & 0.76 & 5.25 & 0.0085 & 252 & 0.34 \\
HunyuanImage-3.0-Instruct & 151.52 & 0.79 & 0.74 & 5.42 & 0.0098 & \textbf{254} & \textbf{0.41} \\
Qwen-Image-Edit-2511 & 171.47 & 0.76 & 0.56 & 4.93 & 0.0129 & 220 & 0.27 \\
OmniGen2 & 160.56 & 0.79 & 0.67 & 4.93 & 0.0130 & 229 & 0.32 \\
\bottomrule
\end{tabular}
\end{table*}

\begin{table*}[t]
\centering
\small
\caption{Holistic metrics on the ablation subset of \textsc{MultiBind} when subject reference images are synthesized by Seedream 4.5. The ``Matched'' metric reports the number of aligned subject slots within the all-model success intersection for this reference-generator setting.}
\label{tab:ablation_holistic_seedream_ref}
\begin{tabular}{lccccccc}
\toprule
Model & FID$\downarrow$ & CLIP-I$\uparrow$ & DINO$\uparrow$ & AES$\uparrow$ & JS$\downarrow$ & Matched$\uparrow$ & Mean IoU$\uparrow$\\
\midrule
Nano Banana Pro & 139.04 & \underline{0.88} & \underline{0.82} & \underline{5.02} & \textbf{0.0068} & \underline{254} & 0.41 \\
GPT-Image-1.5 & \underline{138.84} & 0.85 & 0.79 & \textbf{5.39} & 0.0083 & 250 & 0.38 \\
Seedream 4.5 & \textbf{129.09} & \textbf{0.89} & \textbf{0.84} & 4.58 & \underline{0.0075} & \textbf{256} & \textbf{0.45} \\
HunyuanImage-3.0-Instruct & 139.47 & 0.81 & 0.77 & 4.93 & 0.0102 & 248 & \underline{0.42} \\
Qwen-Image-Edit-2511 & 166.17 & 0.78 & 0.59 & 4.60 & 0.0140 & 210 & 0.24 \\
OmniGen2 & 152.04 & 0.81 & 0.71 & 4.47 & 0.0137 & 209 & 0.26 \\
\bottomrule
\end{tabular}
\end{table*}

Across the three tables, the overall structure is stable. Nano Banana Pro, GPT-Image-1.5, and Seedream 4.5 remain the strongest group on fidelity-oriented metrics, while Qwen-Image-Edit-2511 and OmniGen2 remain clearly behind on nearly all metrics. HunyuanImage-3.0-Instruct again exhibits a distinctive profile: it is usually weaker than the top group on CLIP-I/DINO, but remains competitive or best on Matched and Mean IoU, indicating relatively strong slot alignment even when subject fidelity is less robust.

We also do not observe a uniform same-model bonus. GPT-Image-1.5 does not become consistently better when GPT-generated references are used: its best FID and DINO appear with Seedream-generated references, while its best CLIP-I, Matched, and Mean IoU appear with Nano Banana-generated references. Nano Banana Pro likewise is not uniquely helped by Nano Banana-generated references: its best CLIP-I and JS are obtained with Seedream-generated references, whereas its best FID, DINO, and Mean IoU appear with Nano Banana-generated references. Seedream 4.5 does improve noticeably under Seedream-generated references on FID/CLIP-I/DINO and Mean IoU, but this self-alignment effect is not mirrored as a universal pattern across models and therefore does not support a broad claim that our default Nano Banana-generated references materially favor Nano Banana Pro.

Among the holistic metrics, JS seems slightly more sensitive to the choice of reference generator, but the absolute values remain small. For the strongest models, JS stays roughly in the $0.0068$--$0.0085$ range across all three settings, and even the weaker models remain on the order of $10^{-2}$. Overall, the holistic results are consistent with the pattern-rate analysis below: changing the reference-image generator can move individual numbers, but it does not materially change the benchmark's qualitative conclusions or reveal a systematic unfairness toward models evaluated against references produced by themselves.

\subsection{Pattern Rates}
\label{sec:supp_ablation_reference_generator_pattern}

Tables~\ref{tab:ablation_pattern_rates_banana_ref}--\ref{tab:ablation_pattern_rates_seedream_ref} report the same thresholded subject-level and image-level diagnostics as in Table~4 of the main paper. We keep the thresholds fixed to the main-paper calibration and only change the reference-image generator, so the differences below reflect changes in the synthetic references rather than a redefinition of the metric.
\begin{table*}[!t]
\centering
\small
\caption{Pattern rates (\%) on the ablation subset when subject reference images are synthesized by Nano Banana Pro. We keep the binarization thresholds fixed to the values calibrated in the main paper. Success, Confused, Inconsistent, and Drift are subject-level rates; Swap, Dominance, and Blending are image-level pattern rates. Abbrev.: Suc.=Success, Conf.=Confused, Inc.=Inconsistent, Dom.=Dominance, Bld.=Blending. HunyuanImage-3 denotes HunyuanImage-3.0-Instruct; Qwen-Image-Edit denotes Qwen-Image-Edit-2511.}
\label{tab:ablation_pattern_rates_banana_ref}
\begin{tabular}{@{}llcccc@{}ccc@{}}
\toprule
Dim & Model &
\multicolumn{4}{c}{Subject-level (\%)} &
\multicolumn{3}{c}{Image-level (\%)} \\
\cmidrule(lr){3-6}\cmidrule(lr){7-9}
& &
\multicolumn{1}{c}{Suc.$\uparrow$} &
\multicolumn{1}{c}{Conf.$\downarrow$} &
\multicolumn{1}{c}{Inc.$\downarrow$} &
\multicolumn{1}{c}{Drift$\downarrow$} &
\multicolumn{1}{c}{Swap$\downarrow$} &
\multicolumn{1}{c}{Dom.$\downarrow$} &
\multicolumn{1}{c}{Bld.$\downarrow$} \\
\midrule
\multirow{6}{*}{Face} & \mbox{Nano Banana Pro} & \textbf{85.6} & 11.9 & \textbf{4.7} & \textbf{2.5} & \underline{1.3} & \underline{3.8} & 24.1 \\
 & \mbox{GPT-Image-1.5} & \underline{83.5} & \textbf{10.6} & \underline{7.2} & \underline{5.9} & \textbf{0.0} & \textbf{2.5} & \underline{21.5} \\
 & \mbox{Seedream 4.5} & 61.2 & 32.3 & 11.2 & 6.5 & 3.8 & 10.3 & 51.3 \\
 & \mbox{HunyuanImage-3} & 40.2 & \underline{10.7} & 56.4 & 49.1 & 5.1 & 5.1 & \textbf{11.5} \\
 & \mbox{Qwen-Image-Edit} & 45.6 & 36.3 & 40.4 & 18.1 & 19.2 & 15.4 & 30.8 \\
 & \mbox{OmniGen2} & 45.5 & 33.3 & 42.9 & 21.2 & 21.5 & 11.4 & 27.8 \\
\midrule
\multirow{6}{*}{Appearance} & \mbox{Nano Banana Pro} & \underline{94.9} & \underline{4.7} & \underline{4.3} & \underline{0.4} & \underline{3.4} & \underline{1.1} & \textbf{4.5} \\
 & \mbox{GPT-Image-1.5} & \textbf{95.7} & \textbf{4.3} & \textbf{2.4} & \textbf{0.0} & \textbf{2.3} & \textbf{0.0} & \textbf{4.5} \\
 & \mbox{Seedream 4.5} & 92.9 & 6.3 & 5.2 & 0.8 & \textbf{2.3} & \textbf{0.0} & \underline{9.2} \\
 & \mbox{HunyuanImage-3} & 80.0 & 14.9 & 13.7 & 5.1 & \textbf{2.3} & \textbf{0.0} & 19.3 \\
 & \mbox{Qwen-Image-Edit} & 50.9 & 37.6 & 43.6 & 11.5 & 23.0 & 9.2 & 25.3 \\
 & \mbox{OmniGen2} & 58.7 & 34.9 & 36.7 & 6.4 & 20.5 & 5.7 & 20.5 \\
\midrule
\multirow{6}{*}{Pose} & \mbox{Nano Banana Pro} & \textbf{77.0} & \underline{4.0} & \textbf{20.0} & \textbf{19.0} & \textbf{0.0} & 10.5 & 7.9 \\
 & \mbox{GPT-Image-1.5} & \underline{73.5} & 5.1 & \underline{22.4} & \underline{21.4} & \textbf{0.0} & \textbf{2.8} & 11.1 \\
 & \mbox{Seedream 4.5} & 69.4 & \textbf{3.1} & 28.6 & 27.6 & \textbf{0.0} & 8.3 & 5.6 \\
 & \mbox{HunyuanImage-3} & 66.7 & 4.2 & 32.3 & 29.2 & \textbf{0.0} & \underline{5.3} & \underline{5.3} \\
 & \mbox{Qwen-Image-Edit} & 56.2 & 4.7 & 43.8 & 39.1 & \underline{3.6} & 7.1 & \textbf{3.6} \\
 & \mbox{OmniGen2} & 67.8 & 9.2 & 31.0 & 23.0 & \textbf{0.0} & 18.4 & \underline{5.3} \\
\midrule
\multirow{6}{*}{Expression} & \mbox{Nano Banana Pro} & \textbf{94.4} & \textbf{5.6} & 1.2 & \textbf{0.0} & \underline{1.2} & \underline{1.2} & \textbf{9.3} \\
 & \mbox{GPT-Image-1.5} & \underline{94.0} & \underline{6.0} & \underline{0.4} & \textbf{0.0} & \textbf{0.0} & \underline{1.2} & \underline{10.5} \\
 & \mbox{Seedream 4.5} & 92.7 & 7.3 & 1.2 & \textbf{0.0} & \textbf{0.0} & \underline{1.2} & 15.3 \\
 & \mbox{HunyuanImage-3} & 90.0 & 10.0 & \textbf{0.0} & \textbf{0.0} & \textbf{0.0} & \textbf{0.0} & 20.9 \\
 & \mbox{Qwen-Image-Edit} & 69.2 & 29.9 & 8.4 & 0.9 & \underline{1.2} & 17.6 & 40.0 \\
 & \mbox{OmniGen2} & 66.4 & 33.2 & 2.8 & \underline{0.5} & \textbf{0.0} & 14.0 & 51.2 \\
\bottomrule
\end{tabular}
\end{table*}

\begin{table*}[!t]
\centering
\small
\caption{Pattern rates (\%) on the ablation subset when subject reference images are synthesized by GPT-Image-1.5. We keep the binarization thresholds fixed to the values calibrated in the main paper. Success, Confused, Inconsistent, and Drift are subject-level rates; Swap, Dominance, and Blending are image-level pattern rates. Abbrev.: Suc.=Success, Conf.=Confused, Inc.=Inconsistent, Dom.=Dominance, Bld.=Blending. HunyuanImage-3 denotes HunyuanImage-3.0-Instruct; Qwen-Image-Edit denotes Qwen-Image-Edit-2511.}
\label{tab:ablation_pattern_rates_gpt_ref}
\begin{tabular}{@{}llcccc@{}ccc@{}}
\toprule
Dim & Model &
\multicolumn{4}{c}{Subject-level (\%)} &
\multicolumn{3}{c}{Image-level (\%)} \\
\cmidrule(lr){3-6}\cmidrule(lr){7-9}
& &
\multicolumn{1}{c}{Suc.$\uparrow$} &
\multicolumn{1}{c}{Conf.$\downarrow$} &
\multicolumn{1}{c}{Inc.$\downarrow$} &
\multicolumn{1}{c}{Drift$\downarrow$} &
\multicolumn{1}{c}{Swap$\downarrow$} &
\multicolumn{1}{c}{Dom.$\downarrow$} &
\multicolumn{1}{c}{Bld.$\downarrow$} \\
\midrule
\multirow{6}{*}{Face} & \mbox{Nano Banana Pro} & \textbf{76.2} & 11.9 & \textbf{15.7} & \underline{11.9} & \textbf{0.0} & 6.3 & 20.3 \\
 & \mbox{GPT-Image-1.5} & \underline{73.5} & \textbf{9.4} & 20.1 & 17.1 & \underline{1.3} & \underline{3.8} & \underline{17.7} \\
 & \mbox{Seedream 4.5} & 67.4 & 22.3 & \underline{16.3} & \textbf{10.3} & \underline{1.3} & 7.6 & 36.7 \\
 & \mbox{HunyuanImage-3} & 30.5 & \underline{11.0} & 66.5 & 58.5 & 10.1 & \textbf{1.3} & \textbf{11.4} \\
 & \mbox{Qwen-Image-Edit} & 45.5 & 28.5 & 44.5 & 26.0 & 19.0 & 6.3 & 22.8 \\
 & \mbox{OmniGen2} & 38.1 & 28.6 & 52.4 & 33.3 & 21.5 & \underline{3.8} & 25.3 \\
\midrule
\multirow{6}{*}{Appearance} & \mbox{Nano Banana Pro} & \textbf{96.0} & \textbf{3.6} & \textbf{3.2} & \textbf{0.4} & \textbf{1.1} & \underline{2.3} & \underline{4.5} \\
 & \mbox{GPT-Image-1.5} & \underline{95.3} & \underline{4.3} & \textbf{3.2} & \textbf{0.4} & \textbf{1.1} & \textbf{1.1} & 6.8 \\
 & \mbox{Seedream 4.5} & 95.2 & \textbf{3.6} & \underline{4.4} & \underline{1.2} & 3.4 & \underline{2.3} & \textbf{2.3} \\
 & \mbox{HunyuanImage-3} & 77.2 & 18.5 & 16.1 & 4.3 & \underline{2.3} & \textbf{1.1} & 23.9 \\
 & \mbox{Qwen-Image-Edit} & 55.0 & 32.7 & 39.5 & 12.3 & 22.7 & 5.7 & 21.6 \\
 & \mbox{OmniGen2} & 56.8 & 37.6 & 37.6 & 5.7 & 21.6 & \textbf{1.1} & 26.1 \\
\midrule
\multirow{6}{*}{Pose} & \mbox{Nano Banana Pro} & \textbf{78.2} & \underline{4.0} & \textbf{18.8} & \textbf{17.8} & \textbf{0.0} & 7.9 & 7.9 \\
 & \mbox{GPT-Image-1.5} & \underline{72.7} & \underline{4.0} & \underline{25.3} & 23.2 & \textbf{0.0} & 7.9 & \textbf{5.3} \\
 & \mbox{Seedream 4.5} & 68.6 & \textbf{3.9} & 29.4 & 27.5 & \textbf{0.0} & 7.9 & \textbf{5.3} \\
 & \mbox{HunyuanImage-3} & 68.4 & 4.1 & 29.6 & 27.6 & \underline{2.7} & \textbf{0.0} & \underline{5.4} \\
 & \mbox{Qwen-Image-Edit} & 42.3 & 11.5 & 55.8 & 46.2 & 11.5 & \underline{7.7} & 7.7 \\
 & \mbox{OmniGen2} & 61.5 & 16.7 & 32.3 & \underline{21.9} & \textbf{0.0} & 13.2 & 18.4 \\
\midrule
\multirow{6}{*}{Expression} & \mbox{Nano Banana Pro} & \textbf{94.8} & \underline{5.2} & \textbf{0.0} & \textbf{0.0} & \textbf{0.0} & \underline{2.3} & \textbf{10.5} \\
 & \mbox{GPT-Image-1.5} & \underline{93.6} & 6.0 & \underline{0.4} & \underline{0.4} & \textbf{0.0} & \underline{2.3} & \underline{11.6} \\
 & \mbox{Seedream 4.5} & \textbf{94.8} & \textbf{4.8} & \underline{0.4} & \underline{0.4} & \textbf{0.0} & \textbf{1.2} & \textbf{10.5} \\
 & \mbox{HunyuanImage-3} & 86.8 & 13.2 & 2.4 & \textbf{0.0} & \textbf{0.0} & \textbf{1.2} & 23.3 \\
 & \mbox{Qwen-Image-Edit} & 69.4 & 27.8 & 5.6 & 2.8 & \underline{1.2} & 14.0 & 38.4 \\
 & \mbox{OmniGen2} & 65.8 & 33.8 & 5.3 & \underline{0.4} & \textbf{0.0} & 7.0 & 51.2 \\
\bottomrule
\end{tabular}
\end{table*}

\begin{table*}[!t]
\centering
\small
\caption{Pattern rates (\%) on the ablation subset when subject reference images are synthesized by Seedream 4.5. We keep the binarization thresholds fixed to the values calibrated in the main paper. Success, Confused, Inconsistent, and Drift are subject-level rates; Swap, Dominance, and Blending are image-level pattern rates. Abbrev.: Suc.=Success, Conf.=Confused, Inc.=Inconsistent, Dom.=Dominance, Bld.=Blending. HunyuanImage-3 denotes HunyuanImage-3.0-Instruct; Qwen-Image-Edit denotes Qwen-Image-Edit-2511.}
\label{tab:ablation_pattern_rates_seedream_ref}
\begin{tabular}{@{}llcccc@{}ccc@{}}
\toprule
Dim & Model &
\multicolumn{4}{c}{Subject-level (\%)} &
\multicolumn{3}{c}{Image-level (\%)} \\
\cmidrule(lr){3-6}\cmidrule(lr){7-9}
& &
\multicolumn{1}{c}{Suc.$\uparrow$} &
\multicolumn{1}{c}{Conf.$\downarrow$} &
\multicolumn{1}{c}{Inc.$\downarrow$} &
\multicolumn{1}{c}{Drift$\downarrow$} &
\multicolumn{1}{c}{Swap$\downarrow$} &
\multicolumn{1}{c}{Dom.$\downarrow$} &
\multicolumn{1}{c}{Bld.$\downarrow$} \\
\midrule
\multirow{6}{*}{Face} & \mbox{Nano Banana Pro} & \textbf{80.9} & \textbf{12.8} & \textbf{9.4} & \underline{6.4} & \textbf{2.5} & 7.6 & \underline{24.1} \\
 & \mbox{GPT-Image-1.5} & \underline{74.9} & 14.7 & 13.9 & 10.4 & \underline{3.9} & \textbf{1.3} & 28.6 \\
 & \mbox{Seedream 4.5} & 66.4 & 27.7 & \underline{11.3} & \textbf{5.9} & \textbf{2.5} & 3.8 & 40.5 \\
 & \mbox{HunyuanImage-3} & 33.2 & \underline{14.4} & 63.3 & 52.4 & 7.6 & \underline{2.5} & \textbf{11.4} \\
 & \mbox{Qwen-Image-Edit} & 41.7 & 39.6 & 43.9 & 18.7 & 18.2 & 14.3 & 41.6 \\
 & \mbox{OmniGen2} & 43.5 & 37.7 & 41.4 & 18.8 & 10.3 & 17.9 & 35.9 \\
\midrule
\multirow{6}{*}{Appearance} & \mbox{Nano Banana Pro} & 93.7 & 6.3 & 5.1 & \textbf{0.0} & \underline{4.5} & \textbf{1.1} & \textbf{5.7} \\
 & \mbox{GPT-Image-1.5} & \textbf{94.8} & \textbf{4.8} & \textbf{4.0} & \underline{0.4} & \textbf{2.3} & \underline{1.2} & \underline{5.8} \\
 & \mbox{Seedream 4.5} & \underline{94.5} & \underline{5.1} & \underline{4.3} & \underline{0.4} & \textbf{2.3} & \textbf{1.1} & \textbf{5.7} \\
 & \mbox{HunyuanImage-3} & 73.0 & 21.0 & 21.0 & 6.0 & 5.7 & 4.5 & 23.9 \\
 & \mbox{Qwen-Image-Edit} & 53.8 & 37.1 & 42.4 & 9.0 & 23.3 & 8.1 & 22.1 \\
 & \mbox{OmniGen2} & 53.1 & 42.6 & 38.8 & 4.3 & 20.7 & 10.3 & 29.9 \\
\midrule
\multirow{6}{*}{Pose} & \mbox{Nano Banana Pro} & \textbf{78.2} & \underline{3.0} & \textbf{20.8} & \textbf{18.8} & \textbf{0.0} & \underline{7.9} & \textbf{2.6} \\
 & \mbox{GPT-Image-1.5} & \underline{73.9} & 3.3 & \underline{25.0} & \underline{22.8} & \textbf{0.0} & 11.4 & 2.9 \\
 & \mbox{Seedream 4.5} & 72.5 & \textbf{2.9} & 25.5 & 24.5 & \textbf{0.0} & \textbf{5.3} & 5.3 \\
 & \mbox{HunyuanImage-3} & 71.7 & 5.4 & 26.1 & \underline{22.8} & \underline{2.7} & 10.8 & \underline{2.7} \\
 & \mbox{Qwen-Image-Edit} & 35.1 & 7.0 & 63.2 & 57.9 & 10.3 & 10.3 & 3.4 \\
 & \mbox{OmniGen2} & 61.9 & 9.5 & 35.7 & 28.6 & 2.8 & 19.4 & 8.3 \\
\midrule
\multirow{6}{*}{Expression} & \mbox{Nano Banana Pro} & 93.6 & 6.4 & \textbf{0.8} & \textbf{0.0} & \textbf{0.0} & 2.3 & 12.8 \\
 & \mbox{GPT-Image-1.5} & \underline{94.3} & \underline{5.3} & \textbf{0.8} & \underline{0.4} & \textbf{0.0} & \underline{1.2} & \textbf{10.7} \\
 & \mbox{Seedream 4.5} & \textbf{94.8} & \textbf{5.2} & \textbf{0.8} & \textbf{0.0} & \textbf{0.0} & \textbf{0.0} & \underline{11.6} \\
 & \mbox{HunyuanImage-3} & 83.2 & 16.8 & \underline{1.2} & \textbf{0.0} & \textbf{0.0} & \underline{1.2} & 32.6 \\
 & \mbox{Qwen-Image-Edit} & 64.6 & 33.5 & 7.3 & 1.9 & \underline{1.2} & 20.2 & 50.0 \\
 & \mbox{OmniGen2} & 62.9 & 36.1 & 3.4 & 1.0 & \textbf{0.0} & 15.3 & 51.8 \\
\bottomrule
\end{tabular}
\end{table*}

Across Tables~\ref{tab:ablation_pattern_rates_banana_ref}--\ref{tab:ablation_pattern_rates_seedream_ref}, the coarse model ranking is remarkably stable. Nano Banana Pro, GPT-Image-1.5, and Seedream 4.5 remain the strongest group overall, especially on appearance and expression, while Qwen-Image-Edit and OmniGen2 remain the most confusion-prone models, with large swap, dominance, and blending rates. HunyuanImage-3 continues to occupy a distinct regime with relatively limited confusion in some dimensions but severe self-degradation, especially on face identity.

The same failure patterns also persist across reference generators. Seedream 4.5 remains face-mixing heavy: its face blending stays high under Nano Banana Pro / GPT-Image-1.5 / Seedream 4.5 references (51.3 / 36.7 / 40.5). By contrast, HunyuanImage-3 remains face-drift heavy, with face drift 49.1 / 58.5 / 52.4 and much lower face blending than Seedream 4.5 in all three tables. Qwen-Image-Edit and OmniGen2 remain unstable on both counts, especially on appearance and expression where swap, dominance, and blending all stay high. Pose is still primarily limited by inconsistency/drift rather than confusion for most models, while expression remains a low-drift but non-trivially blending-prone dimension.

Most importantly, we do not observe a systematic same-model advantage. Nano Banana Pro remains top or near-top even when the references are regenerated by GPT-Image-1.5 or Seedream 4.5. GPT-Image-1.5 does not become uniformly better under GPT-generated references; for example, its face success is 73.5 with GPT-generated references, versus 83.5 under Nano Banana Pro-generated references and 74.9 under Seedream-generated references. Seedream 4.5 also does not lose its characteristic face-mixing profile when Seedream-generated references are used. Therefore, although absolute values do shift on this smaller ablation subset, the overall diagnostic picture is very similar across the three reference-generation choices. This suggests that using Nano Banana Pro to synthesize a large portion of the benchmark references does not introduce a material unfairness in the evaluation.

%% file: supplements/more_cases.tex
\clearpage
\newcommand{\maybequalcase}[3]{
\begin{figure*}[!b]
\centering
\IfFileExists{#1}{
  \includegraphics[width=1.0\textwidth]{#1}
}{
  \fbox{
    \parbox[c][0.35\textheight][c]{0.95\textwidth}{\centering Missing figure asset:\\[4pt]\texttt{\detokenize{#1}}}
  }
}
\caption{#2}
\label{#3}
\end{figure*}
}

\maybequalcase{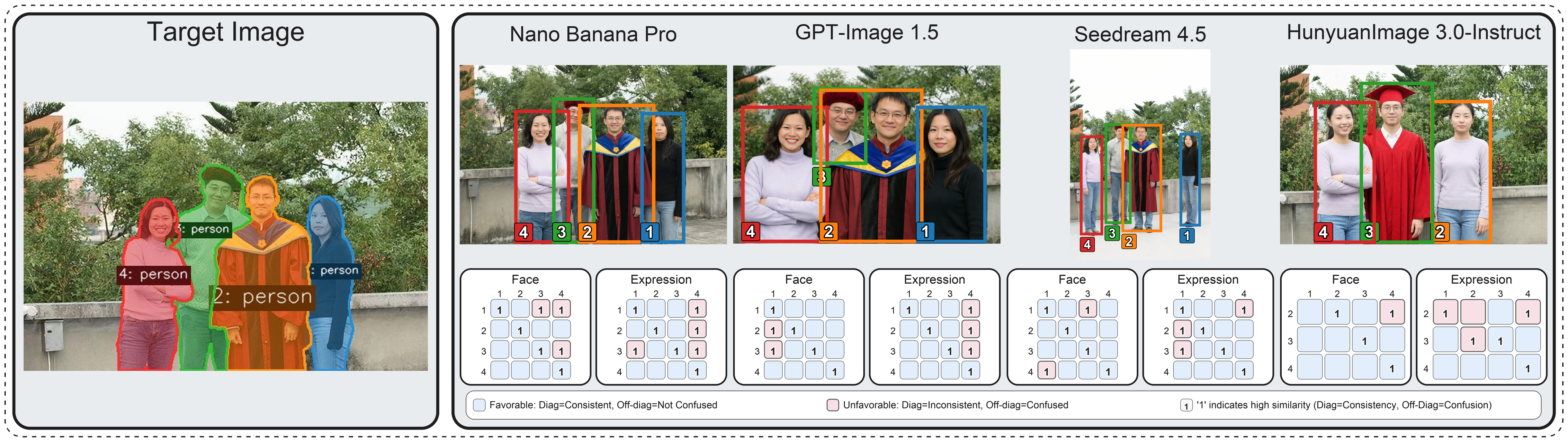}{Additional qualitative example 1.}{fig:more_cases_1}
\maybequalcase{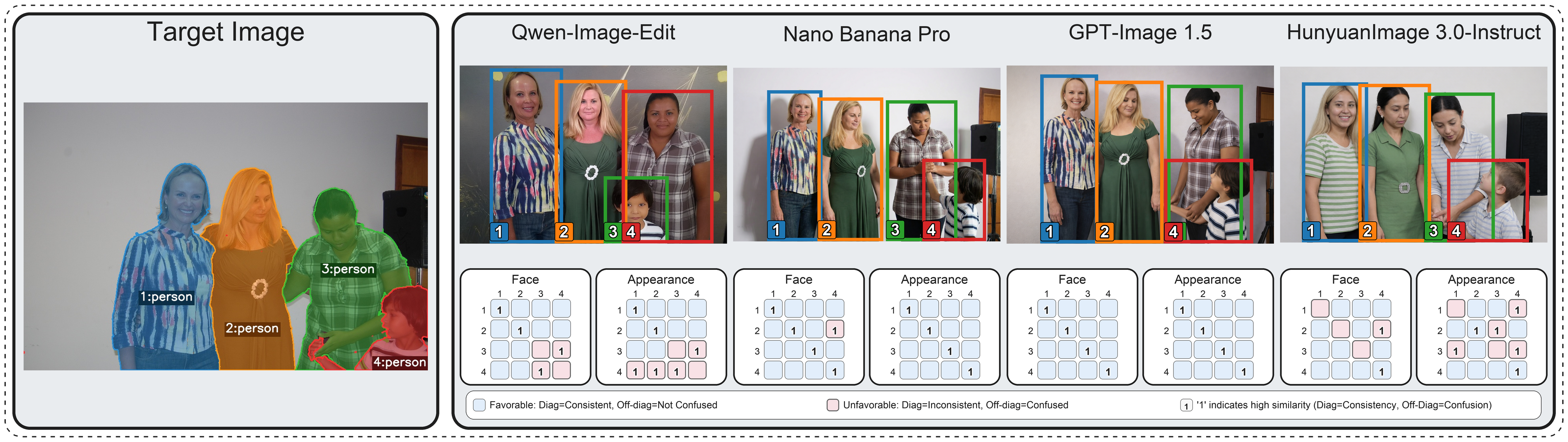}{Additional qualitative example 2.}{fig:more_cases_2}
\maybequalcase{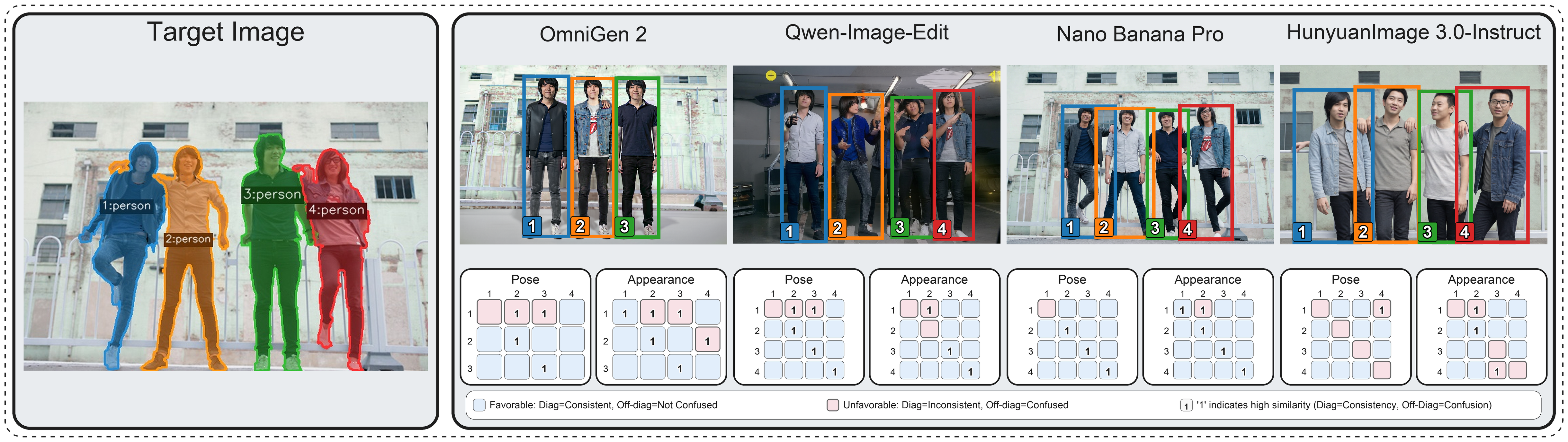}{Additional qualitative example 3.}{fig:more_cases_3}
\maybequalcase{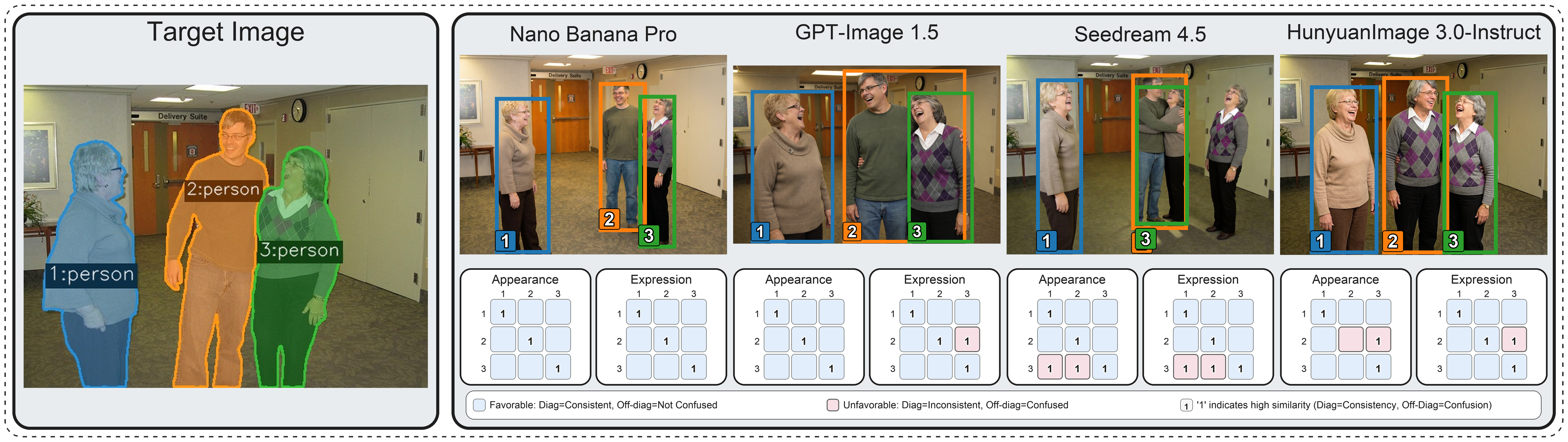}{Additional qualitative example 4.}{fig:more_cases_4}
\maybequalcase{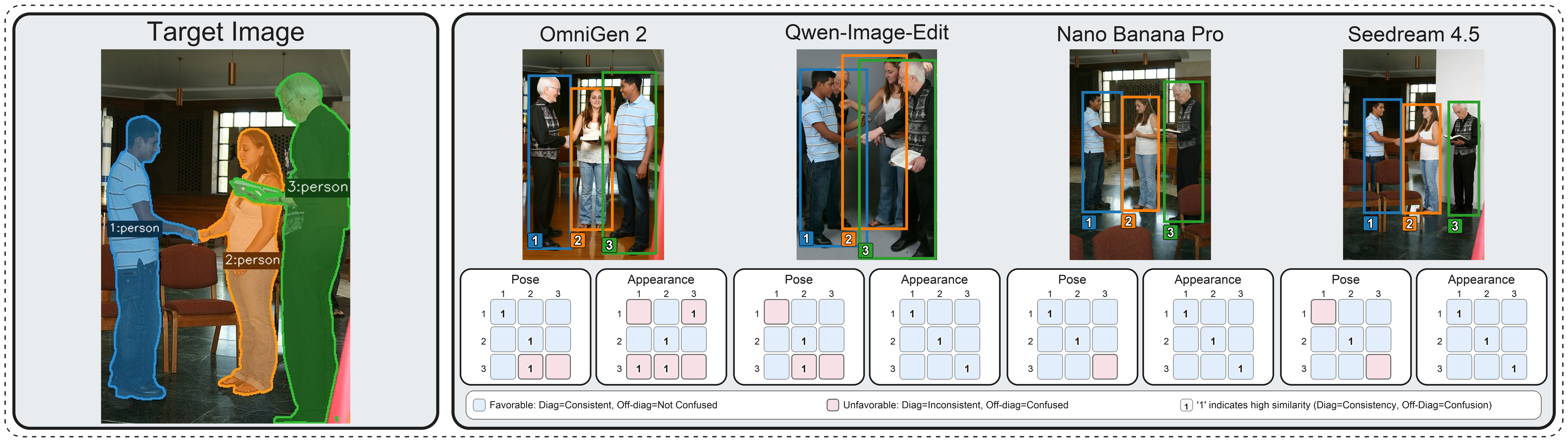}{Additional qualitative example 5.}{fig:more_cases_5}